\documentclass[10pt,twocolumn,letterpaper]{article}

\usepackage[final]{cvpr}      % To produce the REVIEW version
\RequirePackage{paralist}
\RequirePackage{multirow}
\RequirePackage{pdfpages}
\newcommand\blfootnote[1]{%
	\begingroup
	\renewcommand\thefootnote{}\footnote{#1}%
	\addtocounter{footnote}{-1}%
	\endgroup
}
\definecolor{cvprblue}{rgb}{0.21,0.49,0.74}
\RequirePackage[pagebackref,breaklinks,colorlinks,citecolor=cvprblue]{hyperref}
\def\paperID{10225}
\def\confName{CVPR}
\def\confYear{2024}
\usepackage{ulem}
\title{Segment Any Events via Weighted Adaptation of Pivotal Tokens}

\author{Zhiwen Chen$^1$, Zhiyu Zhu$^2$ \footnotemark , Yifan Zhang$^2$, Junhui Hou$^2$, Guangming Shi$^1$, and Jinjian Wu$^1$ \\
	Xidian University$^1$, City University of Hong Kong$^2$\\
	{\tt\footnotesize zhiwen.chen@stu.xidian.edu.cn, zhiyuzhu2-c@my.cityu.edu.hk, yzhang3362-c@my.cityu.edu.hk, }\\ {\tt\footnotesize jh.hou@cityu.edu.hk, gmshi@xidian.edu.cn, jinjian.wu@mail.xidian.edu.cn}
}
\begin{document}
	% \maketitle
	% \begin{figure*}[ht]
	%     \centering
	%     \resizebox{1.0\textwidth}{!}{\includegraphics[trim={0 0 0 0},clip]{24/Figs/Event_Segmentation.png}}
	%     % \vspace{-5cm}
	%     \caption{Illustration of sequence samples from our RGB-Event MOT benchmark. (a) RGB-based MOT results. (b) Our RGB-Event MOT results. (c) Quantitative comparisons of RGB \textit{vs.} RGB-Event methods.}
	%     \label{fig:teaser}
	% \end{figure*}
	% \twocolumn[{
	% \maketitle
	% }]
	
	\twocolumn[{
		\maketitle
		\begin{center}
			\captionsetup{type=figure}
			\resizebox{1.0\textwidth}{!}{\includegraphics[trim={0 0 0 0},clip]{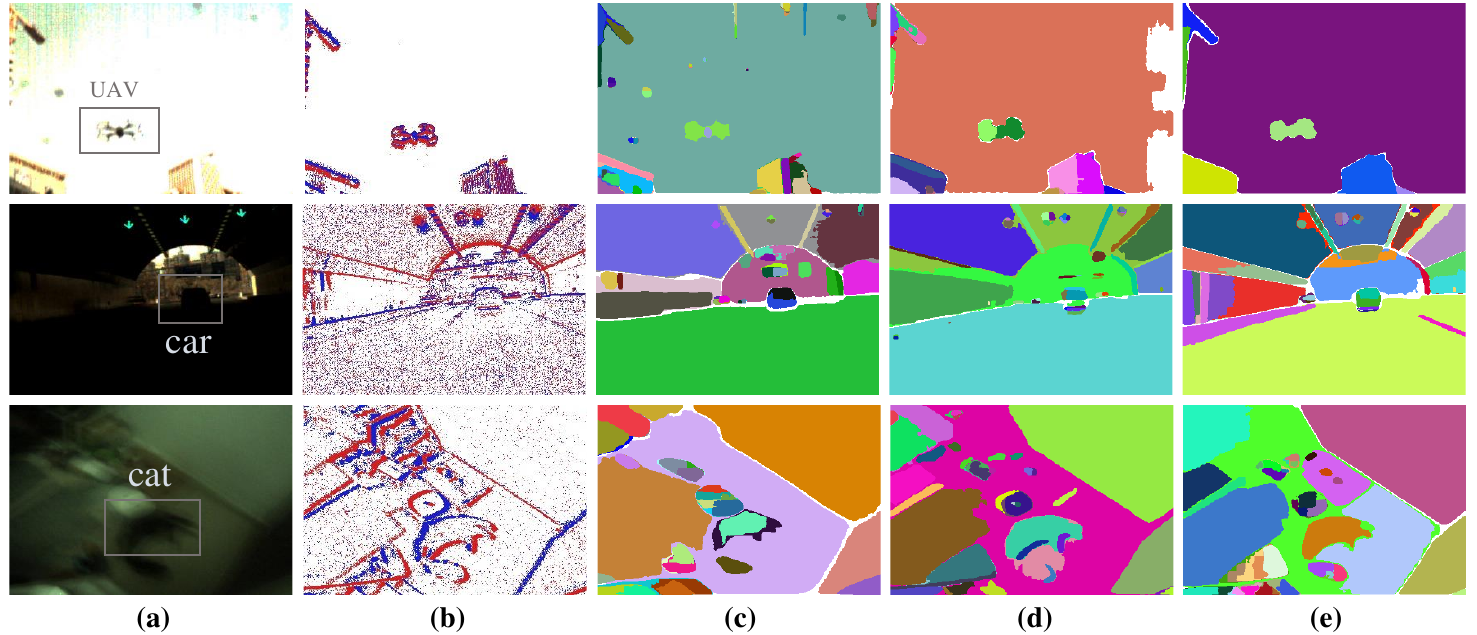}}\vspace{-0.3cm}
			\captionof{figure}{Illustration of object segmentation with different Segment Anything Models (SAMs)-based methods, where (a) RGB images, (b) event data (polarity map for visualization), (c) segmentation results by the original SAM w/ RGB images, (d) the original SAM w/ event data, and (e) our fine-tuned SAM w/ event data. \textbf{Please refer to the supplementary for more demos and applications.}}
			\label{fig:Cover}
			\vspace{-0.1cm}
		\end{center}
	}]
	% \saythanks
	% \setcounter{footnote}{\value{savecntr}}\footnotetext{Corresponding author: Zhiyu Zhu}

	\begin{abstract}
		\blfootnote{$^*$ The first two authors contributed to this paper equally. Corresponding author: Zhiyu Zhu.}
		
		\vspace{-0.6cm}
		In this paper, we delve into the nuanced challenge of tailoring the Segment Anything Models (SAMs) for integration with event data, with the overarching objective of attaining robust and universal object segmentation within the event-centric domain. One pivotal issue at the heart of this endeavor is the precise alignment and calibration of embeddings derived from event-centric data such that they harmoniously coincide with those originating from RGB imagery. Capitalizing on the vast repositories of datasets with paired events and RGB images, our proposition is to harness and extrapolate the profound knowledge encapsulated within the pre-trained SAM framework. As a cornerstone to achieving this, we introduce a multi-scale feature distillation methodology. This methodology rigorously optimizes the alignment of token embeddings originating from event data with their RGB image counterparts, thereby preserving and enhancing the robustness of the overall architecture. Considering the distinct significance that token embeddings from intermediate layers hold for higher-level embeddings, our strategy is centered on accurately calibrating the pivotal token embeddings. This targeted calibration is aimed at effectively managing the discrepancies in high-level embeddings originating from both the event and image domains. Extensive experiments on different datasets demonstrate the effectiveness of the proposed distillation method. Code in \hyperlink{blue}{https://github.com/happychenpipi/EventSAM}.
	\end{abstract}
	
	\vspace{-0.6cm}
	\section{Introduction}
	As the vanguard of sensor technology, event cameras offer a suite of compelling advantages that compensate traditional image sensors. These advantages include, but are not limited to, unparalleled temporal resolution \cite{rebecq2019high,scheerlinck2020fast}, expansive dynamic range \cite{messikommer2022multi,gallego2020event,schiopu2023entropy}, markedly reduced latency \cite{dimitrova2020towards}, and commendable energy efficiency \cite{ramesh2020low}. Current research in event cameras has showcased their vast potential through a multitude of applications, including object classification \cite{deng2020amae}, precise detection \cite{li2022asynchronous,afshar2020event,tomy2022fusing}, tracking \cite{mitrokhin2018event,wang2023visevent}, and segmentation \cite{alonso2019ev,sun2022ess}. Despite these advances, event-based research always faces with the challenge of the limited scale of annotated datasets, which are often constrained in both scale and diversity. This bottleneck severely hinders the ability of algorithms to generalize beyond controlled environments and to process the variable and unpredictable nature of real-world scenes effectively.
	
	In the contemporary landscape of visual perception, there has been a gradual shift, primarily fueled by the advanced capabilities of deep learning architecture techniques, such as transformers. These innovations have yielded a new era of performance in image-based tasks, achieving unprecedented success in object recognition \cite{sironi2018hats}, detection \cite{gehrig2023recurrent}, and segmentation \cite{sun2022ess}. Yet, this progress has predominantly occurred under nearly ideal conditions, which include well-controlled lighting and minimal object movement, creating an observable disconnect from the complex and often less-than-ideal conditions of real-world applications. Traditional image processing techniques frequently decline when faced with challenging environments, such as poorly illuminated areas \cite{wei2018deep} or scenes containing high-speed objects \cite{bochinski2017high}. To bridge this divide, there is a pressing need to combine the unique sensing characteristics of event-based cameras with the extensive repositories of knowledge embedded within large-scale image datasets and the advanced learning capabilities of pre-trained models. Doing so could catalyze the development of more robust and universal visual perception systems that can confidently navigate and interpret complex, dynamic environments, shown as Fig.~\ref{fig:Cover}.
	% Moreover, it could facilitate the translation of this advanced sensory data into actionable insights, further promoting the fields of autonomous navigation, robotics, and real-time monitoring.
	
	Due to the huge distribution gap between the image and event domains, it's not easy for us to directly adapt pre-trained models onto the event data. Fortunately, a considerable number of datasets with paired image and event data have been proposed recently \cite{zhu2018ev,wang2023visevent,tang2022revisiting}. To take advantage of such abundant cross-modal pairs, we propose to transfer the rich knowledge of pre-trained SAM to event domain. Given that the dominant information and weights are in encoder part of SAM, i.e., the ViT backbone \cite{zhang2023faster}, we enforce a calibration of high-level token embeddings from event domain to the image embeddings. Specifically, to avoid the loss of knowledge and well adapt the distribution of image and event data, we mix event and a minimal subset of image tokens together into student-ViT backbone and minimize the gap between the hierarchical embeddings from event and image domains. However, owing to the inherent distinction between the event and image domains, fully eliminating the discrepancies between different modalities is not only intractable but also deemed impossible. Consequently, we put forth a proposition to prioritize network optimization on the intermediate pivotal token embeddings, thereby effectively alleviating the gap that exists within the deeper layers.
	
	The contributions of the proposed method mainly lie in following three-fold:
	\begin{compactitem}
		\item we make the first attempt to adapt the SAMs for event data, which results in event-based universal object segmentation models;
		
		\item  we propose to weight the regularization of the intermediate token embeddings by a approximated significance, thereby facilitating the refinement of the embeddings in the terminal layer; and
		
		% \item we explore different techniques to 
		
		\item we carry extensive experiments to evaluate the effectiveness of the proposed method.
	\end{compactitem}
	% Motivated by aforementioned points, we proposed
	
	% Facing that dilemma, we propose to generalize the
	
	\section{Related Work}
	\label{sec:rw}
	\subsection{Cross-modal Knowledge Distillation and Transfer}
	
	\textbf{Knowledge distillation and transfer of classical neural networks.} Knowledge distillation or transfer is a common tool for researchers to achieve efficient visual recognition and adaptation of different domains or modalities \cite{gou2023hierarchical,gou2021knowledge}. Thoker \textit{et al.} \cite{thoker2019cross} utilized the mutual learning techniques with multiple students to distill knowledge from an image pre-trained recognition network onto human skeleton data for action recognition.  Hu \textit{et al.} \cite{hu2020creating} proposed an unsupervised knowledge distillation framework to calibrate the embedding distribution from achieving cross-modal hashing. Gupta \textit{et al.} \cite{gupta2016cross} made use of the knowledge from the teacher model to train the student model on a new unlabeled input modality, e.g., the paired depth image and optical flow.
	Garcia \textit{et al.} \cite{garcia2018modality} achieved cross-modality distillation with an additional modality of depth image to reconstruct a hallucination stream. Tian \textit{et al.} \cite{tian2019contrastive} applied a contrastive loss to distillate pair-wise relationships across different modalities. Roheda \textit{et al.}~\cite{roheda2018cross} proposed to compensate for the missing modalities and achieve knowledge distillation on available modalities using GANs. Do \textit{et al.}~\cite{do2019compact} utilized a cross-modal knowledge distillation method for a visual question answering method. 
	
	\vspace{0.7em}
	\noindent\textbf{Knowledge distillation of large pre-trained models.} Recently, with the emergence of large pre-trained models, e.g., GPTs~\cite{brown2020language}, llamas~\cite{touvron2023llama}, SAMs~\cite{kirillov2023segment} and stable diffusion models~\cite{rombach2021highresolution}, many works attempt to adapt the pre-trained models with specific tasks on other modalities to make use of their capacities.  Fathullah ~\textit{et al.}~\cite{fathullah2023prompting} proposed to adapt LLMs to speech recognition. Mohit~\textit{et al.}~\cite{sharma2023lossless} introduced a large pre-trained model for robotic manipulation. Chu~\textit{et al.}~\cite{chu2023leveraging} introduced pre-trained LLMs for recommender systems. \cite{tagliabue2023real} adapted language models onto aerial robots. He~\textit{et al.}~\cite{he2021effectiveness} proposed an adapter-based tuning method for the adaptation of large pre-trained LLMs on specific tasks. Mondal~\textit{et al.}~\cite{mondal2023equivariant} proposed to adapt pre-trained SAMs with different input transformations for robust object recognition.
	
	In summary, many delicate and impressive methods have been proposed for knowledge distillation. However, most of them are built on human priors/thoughs, e.g., naturally matching the feature maps from different views or modalities, but with less consideration of the inherent characteristics of neural networks. At the age of Transformer, the network is quite easy to interpret by its inherent attention matrix pattern. Thus, we would further improve those methods via analysis of the inherent demands of neural networks.
	
	\begin{figure*}
		\centering
		\includegraphics[clip, width=0.9\textwidth]{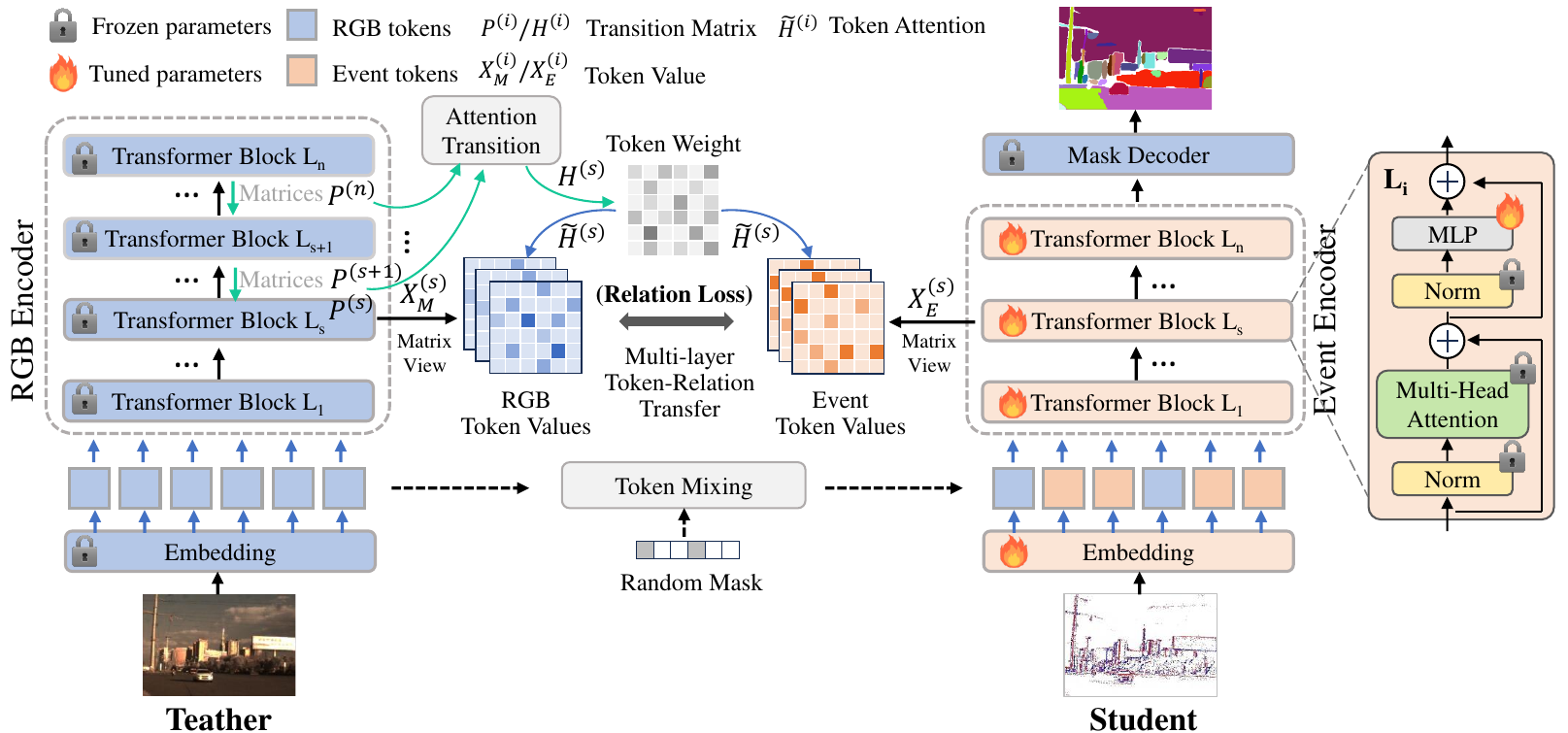}
		\vspace{-0.2cm}
		\caption{Training workflow for knowledge distillation from pre-trained SAMs to the event domain. We employ the original SAMs as a teacher network to derive meaningful semantic features from RGB tokens. Subsequently, we align multi-layer event token embeddings with image embeddings to enable knowledge distillation. Here, we adopt the 3-channel time-voxel image as input, the polarity image only for visualization. The training is facilitated by simultaneously inputting mixed event and RGB tokens into the student-SAMs~(Only event data in the testing phase). To preserve knowledge integrity, we frozen the most of original neural network weights, focusing fine-tuning efforts on the final MLPs in each block. Lastly, we introduce a novel attention-aware embedding weighting strategy to efficiently regularize intermediate-layer token embeddings and significantly improve distillation effectiveness.}
		\label{fig:Framework}
		\vspace{-0.6cm}
	\end{figure*}
	\subsection{Event-based Vision}
	Owing to its innate characteristics of high temporal resolution and dynamic range, event-based vision has been a progressively prevalent subject for research in recent years, e.g., frame reconstruction~\cite{weng2021event,pan2020high,liang2023coherent,zhou2023deblurring,yang2023learning}, interpolation~\cite{tulyakov2021time,tulyakov2022time,sun2023event,kim2023event}, flow estimation~\cite{shiba2022secrets,wan2023rpeflow,wan2022learning,zhu2018ev,gallego2020event,gallego2019focus}, object detection~\cite{mitrokhin2018event,li2022asynchronous,gehrig2023recurrent,schaefer2022aegnn} and tracking~\cite{zhang2022spiking,zhang2021object, wang2023visevent,tang2022revisiting, zhu2022learning,zhu2023cross}. Among them, object segmentation using event data is also one of the most popular research topics for robust object recognition~\cite{stoffregen2019event,sun2022ess}. Specifically, Stoffregen~\textit{et al.}~\cite{stoffregen2019event} proposed to utilize the motion information exhibited in event data to achieve motion segmentation. Mitrokhin~\cite{mitrokhin2020learning} proposed to utilize event surfaces to achieve visual motion segmentation. Alonso~\textit{et al.}~\cite{alonso2019ev} utilized a gray-scale Cityscapes dataset to train a network to label event dataset for training an event-based semantic segmentation network. Zhou~\textit{et al.}~\cite{zhou2021event} proposed a graph-cut based approach to achieve object motion segmentation on an event dataset. Sun~\textit{et al.}~\cite{sun2022ess} introduced a pre-trained E2VID encoder for acquiring rich frame prior for object segmentation. Yang~\textit{et al.}~\cite{yang2020vess} proposed an event-based instance segmentation benchmark. Wang~\textit{et al.}~\cite{wang2021dual,wang2021evdistill} proposed to apply knowledge distillation from image semantic segmentation and E2VID processes for semantic segmentation.
	
	% \textcolor{blue}{Additionally, existing approaches may be broadly classified into two categories: model-based and data-driven. Through describing surrounding environments by a photometric 3D map, Bryner \textit{et al.}~\cite{Bryner2019event} proposed to track the 6-DOF pose of a camera. To capture the spatio-temporal geometry of event data, Mitrokhin \textit{et al.}~\cite{mitrokhin2018event} utilized a parametric model to compensate camera motion.  Based on a pipeline of tracking-learning-detection, Ramesh \textit{et al}.~\cite{Ramesh2018LongtermOT}  proposed an object tracking algorithm for event cameras, which is the first learning-based long-term event tracker. Then, Li \textit{et al.}~\cite{li2019robust} introduced the VGG-Net-16 to encode the appearance of the event-stream object. Inspired by the classic Siamese-matching paradigm, Chae \textit{et al.}~\cite{chae2021siamevent} presented to track objects via learning an edge-aware similarity in the event domain. Recently, Zhang \textit{et al.}~\cite{zhulearning}, introduced a spiking transformer for encoding spatio-temporal information of object tracking. Moreover, ZHU \textit{et al.}~\cite{zhulearning} proposed to utilize inherent motion information of event data to achieve effective object tracking. To summarize, although there are some promising studies that provide directive insights for event-based tracking, a limited number of works have sought to find complementary information from RGB data, e.g., semantic information.}
	
	In summary, although different methods have been proposed to explore and utilize the characteristics of event data, most current research of event cameras is limited to a specific field or dataset, which is quite contradictory with current trends in a universal and unified paradigm for visual recognition with large pre-trained models. However, it's quite effort-consuming to construct large-scale annotated dataset and train the event-based visual perception model from scratch. Thus, utilizing the pre-trained model the image domain to boost performance with event-based vision is a promising approach for building a universal event-based visual perceptual model.
	
	\subsection{Object Segmentation \& SAMs}
	% As one of the most fundamental tasks in computer vision, object segmentation aims to generate pixel-level masks of a target object. Depending on different level annotation masks, it usually be divided into semantic~\cite{long2015fully,noh2015learning}, instance~\cite{liu2018path,lee2020centermask}, panoptic~\cite{kirillov2019panoptic,xiong2019upsnet} segmentation etc. Recently, boosted by massive and finely annotated training data, advanced transformer architecture and delicate task setting definitions, SAMs~\cite{kirillov2023segment} achieve universal object segmentation, with strong zero-shot generalization ability, of a give image with different prompts, e.g., points, bounding boxes or free-form texts. 
	
	In the realm of computer vision, object segmentation is one of cornerstone tasks, focusing on generating pixel-level masks that precisely delineate targeted objects within images~\cite{garcia2017review, asgari2021deep, garcia2018survey}. This intricate process is traditionally segmented into several sub-disciplines based on the nature of mask annotations. Semantic segmentation, as outlined in foundational works~\cite{long2015fully,noh2015learning,yu2018bisenet}, concentrates on classifying each pixel into a fixed set of categories without differentiating between individual object instances. Conversely, instance segmentation~\cite{liu2018path,lee2020centermask,bolya2019yolact}, not only categorizes pixels but also separates different instances of the same category. Panoptic segmentation, a term coined and explored in studies~\cite{kirillov2019panoptic,xiong2019upsnet,cheng2020panoptic}, amalgamates the principles of semantic and instance segmentation to provide a holistic view of scene parsing, distinguishing countable objects.
	
	% The recent paradigm shift in object segmentation can be largely attributed to the fusion of expansive, finely annotated training datasets, which provide a rich tapestry of visual information and contextual nuances. Coupled with this is the emergence of advanced transformer-based architectures, which have revolutionized the field with their ability to capture long-range dependencies and intricate pattern recognitions in images. These architectures are adept at handling the complexity and variability inherent in real-world images.
	
	% Additionally, there has been a meticulous refinement in the definitions and settings of segmentation tasks. This has led to more tailored and precise approaches that cater to specific requirements of diverse applications, ranging from autonomous vehicles to medical image analysis.
	
	Recently, a breakthrough in this domain is the development of SAMs~\cite{kirillov2023segment}. SAMs stand out for their universal object segmentation capability, which is underpinned by a formidable zero-shot generalization ability. This means SAMs can effectively segment objects in images they have never encountered during training. Moreover, SAMs are designed to adapt at interpreting a variety of input prompts, including points, bounding boxes, or free-form texts. 
	% This flexibility allows SAMs to be applied in a wide range of contexts, from automatic image editing to real-time object recognition in dynamic environments.
	
	% To have a comprehensive review of SAM, we give the detailed configuration of it. SAM consists of a MAE~\cite{} based 
	
	\section{Proposed Method}
	
	% To transfer knowledge from pre-trained large vision models, e.g., SAMs, learning similarly distributed high-level embedding from event and image domain is the most essential issue. We start from the architecture of neural network. Generally, the object segmentation networks usually consist of a feature embedding network (encoder) also with a light-weight object information regression network (decoder). 
	% Thus, the most tough issue of this problem is to get well alignment of the feature map from RGB and Event domains. To get detailed investigation of this kind of problem, we first formulate the feature embedding process as,
	The cornerstone of leveraging the capabilities of pre-trained expansive vision models, e.g., SAMs, lies in the effective transfer and distillation of knowledge through aligning high-level embeddings across both event and RGB domains. This endeavor initiates an in-depth analysis of the underlying neural network architecture, which is fundamental to the process of object segmentation. SAMs comprise several components: a feature embedding network, or encoder, designed to extract and process complex visual information, coupled with a streamlined object information regression network, or decoder, aimed at precise pixel-level object classification. To delve into the intricacies of this challenge, it is imperative first to establish a formulation  for the feature embedding process as
	\vspace{-0.2cm}
	\begin{equation}
		\mathcal{F}(\textbf{W}_d,X_d^{in}) = \textbf{X}_{d}^{out}, 
		% \nonumber
	\end{equation}
	\noindent where $d\in \{M,E\}$ indicates the modality of image ($M$) or event data ($E$), $\mathcal{F}(\textbf{W}_M,\cdot)$ (resp, $\mathcal{F}(\textbf{W}_E,\cdot)$) indicates the network with the weights $\textbf{W}_M$ (resp, $\textbf{W}_M$) for feature embedding on RGB images (resp, events), $\textbf{X}_{d}^{out} = \{\textbf{X}_d^{(s_1)}, \cdots, \textbf{X}_d^{(s_n)}\}$ represents a set of multi-layer token embeddings. Thus, this knowledge distillation task is to minimize the distribution gap between $\textbf{X}_{M}^{out}$ and $\textbf{X}_{E}^{out}$. Moreover, due to the fact that the embedding backbone, i.e., ViT, is a plain architecture and only the last layer feature map is fed for object segmentation, the similarity between embedding $\textbf{X}_M^{(s_n)}$ and $\textbf{X}_E^{(s_n)}$ is most pivotal issue.
	
	To adapt the embedding in the event domain and minimize the discrepancy between the RGB and event, we propose the following two regularization strategies.
	
	\vspace{0.5em}
	\noindent  \textbf{1}) We first feed mixed tokens from both modalities to facilitate network training and then extract multi-layer feature maps to fine-tune the model to compensate for the large distribution gaps between the RGB and event images.
	% \st{to minimize the distribution gaps}. 
	
	\noindent \textbf{2}) Due to the distinctive sensing patterns of event and RGB cameras, there are inherent differences in the information captured by image and event data.  \textbf{While we aim to minimize the distribution gaps between these modalities, it is important to acknowledge that completely eliminating such discrepancies is fundamentally impossible.}  Therefore, it is paramount to ascertain the relative importance of individual token embeddings and apply regularizations based on their significance. Previous methods~\cite{selvaraju2017grad,zhou2016learning} have demonstrated that network gradients can serve as indicators of the network's focus. However, calculating the gradient on teacher network would bring an inevitable huge computational burden. Considering the large size of model, it would make the situation even worse.  Fortunately, given the inherent self-attention mechanism of transformers, the focus of network is explicitly encoded as the magnitude of attention values. Thus, we further quantify such a correlation as the self-attention matrix. Then, a weighted distillation algorithm is proposed to focus the training on the pivotal token embeddings. 
	
	% We will illustrate the detailed contents of aforementioned two points in the next sections.
	
	\subsection{Cross-modal Distillation with Mixed Inputs}
	\label{sec:MixedInput}
	As shown in Fig.~\ref{fig:Framework}, during training, we adopt following two simple yet effective ways to minimize the distribution gap between $\textbf{X}_{M}^{out}$ and $\textbf{X}_{E}^{out}$ and facilitate network training. 
	
	% \noindent\textbf{Event Representation.} 
	% Event cameras~\cite{brandli2014240} asynchronously capture pixel-level log-intensity change for triggering events. To make events compatible with the images, following the setting in~\cite{zhu2019unsupervised}, we aggregate events between consecutive frames into a three-dimensional event volume $X_E^{in} \in \mathbb{R}^{ H \times W \times B}$. In our experiment, we set $B = 3$.
	
	\noindent\textbf{Modification of Inputs.} 
	% \st{To minimize the huge distribution gap between the RGB image and event data,} 
	Inspired by current success of mask modeling~\cite{he2022masked,bachmann2022multimae}, we randomly replace a small number of event embeddings with images. We expect that it would help retain the original knowledge and facilitate network training.
	
	\noindent\textbf{Trainable Weights.} Moreover, motivated by the fact that the strong generalization and zero-shot ability of SAMs may stem from their large-scale training dataset (which contains $11M$ images and $1~billion$ masks), we do not change the model structure to reuse the pre-trained model. Furthermore, instead of retraining the whole model, we only fine-tune the several MLPs in network to avoid over-fitting and preserve the intrinsically learned patterns.
	
	Based on the aforementioned techniques, a naïve implementation of event-centric SAM could be trained. In the next section, we will discuss a precise embedding regularization via weighting by intermediate correlations. 
	
	% $\mathcal{L} =\sum_{i=1}^N \alpha_i||X_M^{(i)}-X_E^{(i)}||_1$
	
	\begin{figure}
		\centering
		\includegraphics[clip, width=0.48\textwidth]{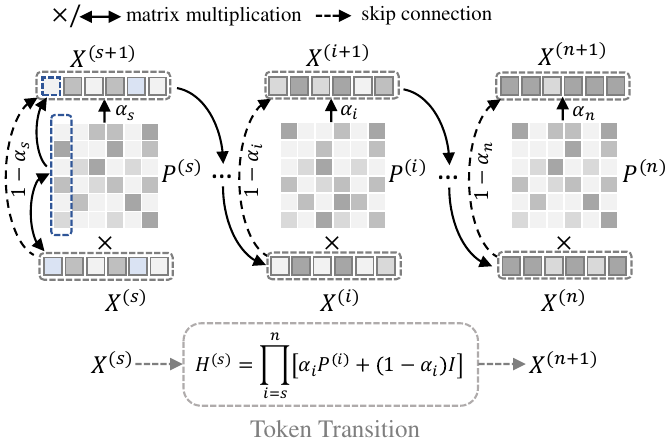}
		\vspace{-0.3cm}
		\caption{The flow of information from specific layer token embeddings $\mathbf{X}^{(s)}$ to higher level token embeddings $\mathbf{X}^{(n+1)}$ in self-attention layers, with gray shades of attention ($\mathbf{P^{(i)}}$) representing higher values and more intensified information transfer. The illustration reveals that across multiple self-attention layers, information originating from a minimal yet crucial set of token embeddings (indicated by the two gray ones on the left inputs $\mathbf{X}^{(s)}$) can significantly influence the embeddings in deeper layers $\mathbf{X}^{(n+1)}$. Consequently, by regulating the features in the initial layers, our approach aims to direct the network's focus towards reducing the disparity between these pivotal embeddings to narrow the gap of embeddings from different modalities.}
		% \textcolor{blue}{Change a color.}
		% \caption{Illustration of information propagation between input and output tokens within self-attention layers, where the darker attention indicates for higher values with strong information transition.  It shows that through transition by multiple self-attention layers, \textbf{information from a minimal set of pivotal tokens (two blue ones in the left) at shallow layer may dominate embeddings at deeper layer}. Thus, through the regularization of shallow layer features, we enforce the network to focus on minimizing discrepancy between those pivotal tokens. }
		% \textcolor{blue}{figure needs to be updated.}
		\label{fig:Attention}
		\vspace{-0.5cm}
	\end{figure}
	
	\subsection{Correlation-aware Weighted Token Distillation}
	
	% To enhance the regularization of intermediate token embeddings and diminish the discrepancy in the terminal layer's outputs of different modalities, we advocate for a strategy that scales the regularization of these intermediate token embeddings according to their pivotal role in influencing the ultimate outputs.
	Initiating from an elemental Vision Transformer (ViT) layer, we quantify the salience of each embedding by treating it as the fundamental unit of message propagation and focusing on the most important inter-embedding information flow. As depicted in the referenced Fig.~\ref{fig:Attention}, we detailedly illustrate the self-attention operation and residuals aggregation in a ViT layer, which facilitate information transition at the inter-embedding level.
	
	We denote $\mathbf{P}^{(i)}\in \mathbb{R}^{k_i \times k_{i+1}}$ as a transition matrix, averaging of multi-head attention on the head dimension, from the $i^{th}$ layer embeddings $\mathbf{X}^{(i)}\in \mathbb{R}^{k_i \times c}$ to its subsequent layer embeddings $\mathbf{X}^{(i+1)}\in \mathbb{R}^{k_{i+1} \times c}$, alongside an identity matrix $\mathbf{I}\in \mathbb{R}^{k_i \times k_{i+1}}$ representing residual aggregation ($k_i \equiv k_{i+1}$  for a plain ViT). The multi-layer inter-embedding transition is then mathematically formulated as 
	% \textcolor{blue}{(there is different with fig.3, we will make a final revision of this point)}
	
	\vspace{-0.3cm}
	\begin{equation}
		\quad \mathbf{H}^{(s)}= \prod_{i=s}^{n} [\alpha_i\mathbf{P}^{(i)} + (1-\alpha_i)\mathbf{I}], 
		% \nonumber
		\vspace{-0.1cm}
	\end{equation}
	\noindent where $\mathbf{H}^{(s)}\in \mathbb{R}^{k_s \times k_{n+1} }$ symbolizes a comprehensive information transition matrix from a specific $s^{th}$ layer to the terminal $n^{th}$ layer. Moreover, the scalar $\alpha_i$ signifies the scaling influence of the normalization layers and MLPs. This results in an alteration of the information ratio coming from the attention mechanism and the residual connections. Due to the fact that it's quite time-consuming and even intractable to properly estimate all the $\alpha_i$, we propose to simplify it by the assumption that the embeddings transition process of multi-layer self-attention mechanism is actually a Markov Chain with a stationary distribution~\cite{paige1975computation}. Thus, we could approximate $\mathbf{H}^{(s)} \approx \widehat{\mathbf{H}}^{(s)}= \beta\prod_{i=s}^{n} \mathbf{P}^{(i)} + (1-\beta)\mathbf{I}$ (Please refer to the supplementary for more analysis). The token-specific transformer's interests/attention from the $s^{th}$ layer, denoted as $\widetilde{\mathbf{H}}^{(s)}\in \mathbb{R}^{k_s}$, is further gauged by multiplying the transition matrix $\mathbf{H}^{(s)}$ with the significance of the final layer's output $\mathbf{e}^T$, formulated as
	
	\vspace{-0.3cm}
	\begin{equation}
		\widetilde{\mathbf{H}}^{(s)} = \widehat{\mathbf{H}}^{(s)} \times \mathbf{e}^T, 
		% \nonumber
	\end{equation}
	\vspace{-0.3cm}
	
	\noindent where we empirically set $\mathbf{e}$ as a vector of ones, underpinning the premise that the embeddings in the last layer uniformly contribute towards the regression of pixel-level object categories. \textit{We calculate the $\widetilde{\mathbf{H}}^{(s)}$ by the attention matrices from the teacher. Please refer to the $2^{nd}$ ablation study in Sec.~\ref{Sec:Ablation} and Table~\ref{table:AblationAttn}.}
	
	\noindent \textbf{Statement of novelty.} Note that some significant advancements have been made in the field of event-based segmentation~\cite{wang2021dual, sun2022ess, wang2021evdistill}. They usually take advantage of image knowledge from an E2VID encoder for object segmentation. In the era of large-scale models, leveraging the original architecture and weights is optimal to maintain the robust zero-shot generalization capabilities inherent in pre-trained backbones~\cite{hu2021lora}. Affinity graph distillation~\cite{wang2021evdistill} presents a compelling approach to guide student networks in acquiring structurally analogous knowledge from their teacher counterparts. The proposed approach, however, emphasizes intensifying the regulation of pivotal proceeding tokens to minimize the distribution gap of feature maps at the terminal layers, implying that the network might permit and manage minor discrepancies in feature maps in less critical regions. Moreover, we avoid the back-propagation of gradients in teacher network by utilizing the off-the-shelf attention matrix in the forward process to approximate the cross-layer embedding correlations, thereby streamlining the process without sacrificing learning efficacy.  
	
	% Considering ViT are stack of transformer blocks, it's nature to decouple them into specific self-attention layers and investigate the .
	% a detailed investigation of a information flow between one token to another for illustration of .
	% To enforce the network minimize the discrepancy between the most important tokens, we quantify the correlation between tokens from each layers via attention values. We 
	
	% part of the network. We investigate the significance of each tokens on different layers, we analysis the correlation between
	
	% \begin{figure}
	%   \centering
	%     \includegraphics[clip, width=0.22\textwidth]{24/Figs/Segmentation_Metric.PNG}
	%   \caption{Illustration of ground truth mask and predicted mask of an instance. Pairs of masks have the largest IoU and are therefore matched. We show how the matched masks are partitioned into TP, FN, and FP.} 
	%     \label{fig:Metric}
	% \end{figure}
	
	\begin{figure*}
		\resizebox{1.0\textwidth}{!}{\includegraphics[trim={0 0 0 0},clip]{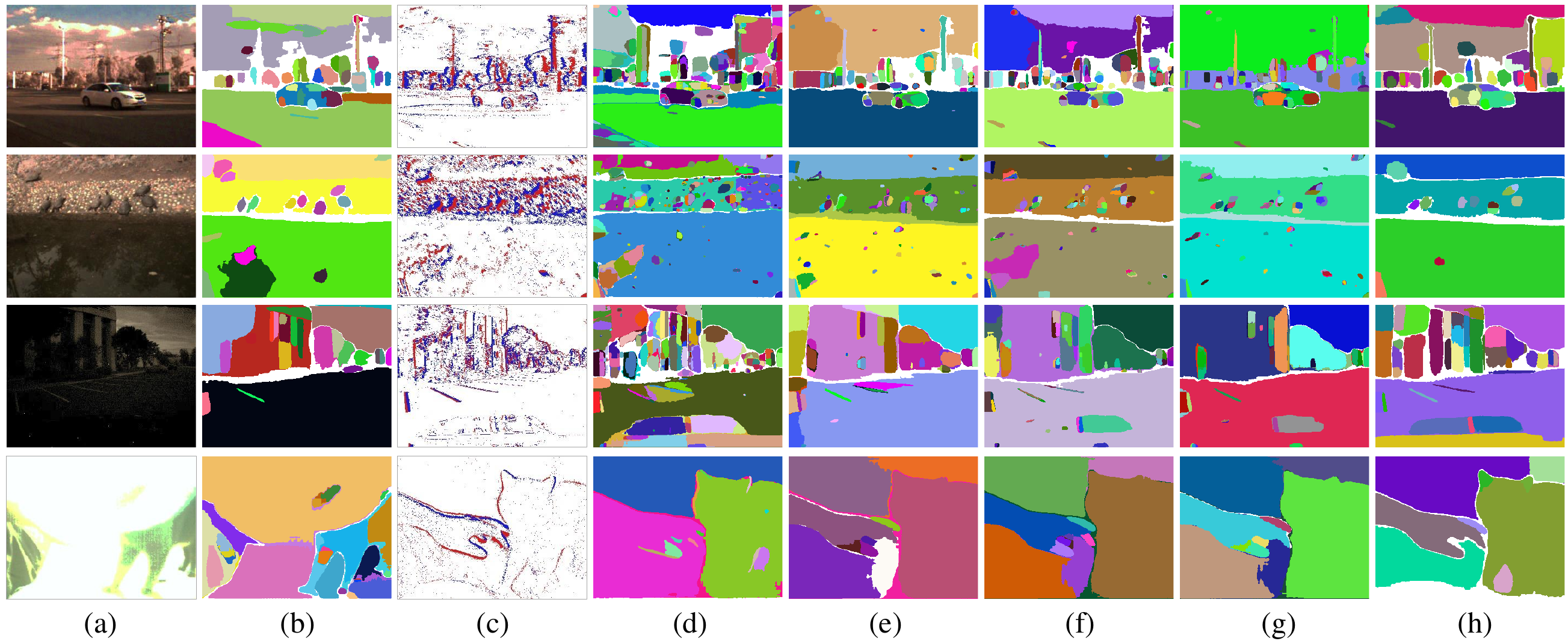}}
		\caption{Visual comparison of different methods on RGBE-SEG (the first two rows), MVSEC (the third row) and the degradation scene (the last row) . (a) RGB image, (b) SAM with image, (c) Event data, (d) E2VID~\cite{rebecq2019high} with event data, (e) NGA~\cite{hu2020learning} with event data, (f) DTL~\cite{wang2021dual} with event data, (g) ESS~\cite{sun2022ess} with event data, and (h) Our method with event data.}
		\label{fig:visual}
		\vspace{-0.2cm}
	\end{figure*}

	\begin{table*}
		\caption{Comparison of the segmentation performance between our method and other representative approaches based on RGBE-SEG dataset. For all metrics, the higher, the better. The best and second-best results under each metric are highlighted in \textbf{bold} and \uline{underlined}, respectively. Note that the mIOU is the most essential metric and mP (mR) cannot comprehensively evaluate the performance.}
		\caption{Comparison of the segmentation performance between our method and other representative approaches based on MVSEC dataset.}
		\label{table:RGBE-SEG}
		\centering
		\footnotesize
		\renewcommand{\arraystretch}{1.2}
		\setlength{\tabcolsep}{1.3mm}
		\begin{tabular}{c| c c c c| c c c c| c c c c| c c c c}
			\toprule[1.2pt]
			\multirow{2}{*}{Method}  &\multicolumn{4}{c|}{Easy} &\multicolumn{4}{c|}{Medium} &\multicolumn{4}{c|}{Hard} &\multicolumn{4}{c}{All}\\
			\cline{2-17}
			\multicolumn{1}{c|}{} &mP & mR &mIoU &aIoU &mP & mR &mIoU &aIoU &mP & mR &mIoU &aIoU &mP & mR &mIoU &aIoU  \\
			\hline
			SAM~\cite{kirillov2023segment}  &0.52   &\textbf{0.75}  &0.39	 &0.58  &0.38  &\underline{0.71}  &0.25	 &0.41    
			&0.26   &0.74  &0.15	 &0.29  &0.39  &\underline{0.73}  &0.26	 &0.43 \\
			E2VID~\cite{rebecq2019high}    &0.62   &0.60  &0.38	 &0.54  &0.55  &0.58  &0.32	 &0.44 &0.46   &0.64  &0.26	 &0.36  &0.54  &0.61  &0.32	    &0.45   \\
			ETNet~\cite{weng2021event}    &\underline{0.68}   &0.56   &\underline{0.40}	 &0.51  &\textbf{0.64}  &0.52  &\underline{0.35}	 &0.42  &\textbf{0.60}   &0.52    &\underline{0.31}  &0.35  &\textbf{0.63}  &0.53   &\underline{0.35}  &0.42   \\
			NGA~\cite{hu2020learning}    &0.56   &0.72  &\underline{0.40}	 &0.58  &0.44  &0.70  &0.29	 &0.45  &0.34  &0.73  &0.21	 &0.35  &0.45  &0.71  &0.30	 &0.45   \\
			LFI~\cite{deng2021learning} &0.56   &0.71  &0.38 &0.56  &0.42  &\underline{0.71}  &0.26	 &0.44 &0.31   &\textbf{0.76}  &0.18	 &0.33  &0.42  &0.72  &0.27	    &0.44  \\ 
			DTL~\cite{wang2021dual} &0.57   &0.71  &\underline{0.40} &\underline{0.59}  &0.45  &0.70  &0.28	 &\underline{0.46} &0.32   &\underline{0.75}  &0.20	 &\underline{0.35}  &0.44  &0.71  &0.29	    &\underline{0.46}  \\ 
			EVDistill~\cite{wang2021evdistill} &\textbf{0.72}  &0.46  &0.32	 &0.40  &\underline{0.64}  &0.46  &0.27	 &0.34  &\underline{0.53}  &0.51  &0.23	 &0.30  &\underline{0.63}  &0.47  &0.27	 &0.34    \\ 
			ESS~\cite{sun2022ess}  &0.55   &\underline{0.73}  &\underline{0.40} &0.57  &0.41  &\textbf{0.73}  &0.27	 &0.45 &0.29   &0.77  &0.18	 &0.32  &0.42  &\textbf{0.74}  &0.28	    &0.45   \\
			\hline
			Ours    &0.66   &\underline{0.73}  &\textbf{0.49} &\textbf{0.65}  &0.61  &0.69  &\textbf{0.40}	 &\textbf{0.54} &0.50   &0.72  &\textbf{0.34}	 &\textbf{0.47}  &0.59  &0.71  &\textbf{0.41}	 &\textbf{0.55}    \\
			\bottomrule[1.2pt]
		\end{tabular}
		\label{Table1}
	\end{table*}
	
	\subsection{Training Objectives}
	
	Combining the aforementioned multi-layer training strategy with the correlation-aware weighting scheme of pivotal token embeddings, we train the network with the following objective function as
	\vspace{-0.2cm}
	\begin{equation}
		\mathcal{L} = \sum_{s_i \in S} \gamma_{i} \mathcal{L}_{s_i}, \quad
		\mathcal{L}_{s_i} = \| \widetilde{\mathbf{H}}^{(s_i)} \times  (\mathbf{X}_M^{(s_i)} - \mathbf{X}_E^{(s_i)})\|_1,
	\end{equation}
	\vspace{-0.3cm}
	
	\noindent where $\gamma_{i}$ denotes a specific weight scalar for $s_i^{th}$ layer embeddings, we broadcast $\widetilde{\mathbf{H}}^{(s_i)}$ to fit the channel dimension of embeddings. Empirically, we regularize four layers of embeddings with a corresponding depth of $\{s_i\}=\{0, 3, 6, 9, 12\}$, where $0$ corresponds to the resulting token embeddings of the patch embedding layer. We do not weight the regularization on the patch embedding layer to learn a robust embedding process. In our experiment, we empirically set $\beta$ as 0.5 and $\gamma_{i}$ as $\{0.1,0.4,0.7,1.0\}$ for these four layers. Note that the segment results of SAM may be dependent upon the distinct prompts employed within the decoder module. Thus, inspired by \cite{zhang2023faster}, we only close the discrepancies between the embeddings to enable an efficient adaptation process and preserve the inherent generalization ability of the decoder.
	
	\section{Experiment}
	\textbf{Datasets.} We collected a large-scale RGB-Event dataset, from current available pixel-level aligned datasets, \textit{i.e.}, VisEvent~\cite{wang2023visevent} and COESOT \cite{tang2022revisiting}. Both of them were captured by the DAVIS346 camera~\cite{brandli2014240}. The DAVIS346 equips a 346$\times$260 pixels active sensor and a dynamic vision sensor, which can acquire aligned images and events. These samples consisted of diverse indoor and outdoor scenarios. For effective knowledge transfer, we only collected image-event pairs without degradation (e.g., no low dynamic range and motion blur) and removed the duplicate samples. After integrating and cleaning, we extended the two datasets to a segmentation dataset, \textbf{RGBE-SEG}. The RGBE-SEG included 65,957 image-event pairs, 64,957 for training and 1,000 for testing. The test set contained 38,760 masks, and we artificially divided it into easy, medium, and hard subsets based on the complexity of scenarios. All ground truth masks were generated by images and the well-trained SAM~\cite{kirillov2023segment}. To further explore the zero-shot performance of our method, we showed more evaluation results on \textbf{MVSEC} dataset~(500 image-event pairs each in "indoor\underline{~}flying1" and "outdoor\underline{~}day2" sequences)~\cite{zhu2018multivehicle}, containing 54,600 masks. 
	
	\vspace{0.5em}
	\noindent \textbf{Evaluation Metrics.} For each instance-level ground truth mask, we define the predicted mask with the maximum intersection over union as its matched mask. This matching splits the pair masks into three sets: true positives (TP), false positives (FP), and false negatives (FN), representing matched segments, unmatched predicted segments, and unmatched ground truth segments, respectively. Based on the split segments, we calculate the precision (P), recall (R) and intersection over union (IoU) to evaluate the segmentation quality. We calculate these metrics for each instance independently and average over instances, obtaining mP, mR, and mIoU. In addition, we introduce the area-weighted intersection-over-union (aIoU). mIoU focuses on the details of scenarios and gives equal treatment to each instance. Instead, aIoU cares about the overall mask quality. These four metrics collaborate to describe the segmentation quality.
	
	\noindent \textbf{Implementation details.} We set the input image and event frame sizes to 512$\times$512. We directly adopted the base vision transformer (ViT-B)~\cite{dosovitskiy2020image} with pre-trained weights as our dual-modal encoders. During training, we used the Adam optimizer to train the event encoder for 5 epochs, 13,500 iterations, with a batch size of 24. We set the initial learning rate as 2e-4. The learning rate is adjusted by a decay scheduler, which is scaled by 0.9 on the $4^{th}$ epoch. All networks are implemented in PyTorch and run on a computer equipped with 4 GPUs (GeForce RTX 3090). 
	
	\noindent \textbf{Comparing methods.} To fully investigate the capacity of the proposed method, we investigate   the performance of different SOTA event-based segmentation methods. Note that the SAM could predict much more precise masks than other segmentation models. Thus, it's unfair to directly adopt those pre-trained models for comparison. Instead, we retrain different distillation and adaptation methods to adapt SAMs. Generally, they could be divided into three categories. The first directly applied frame reconstruction algorithm, e.g., E2VID~\cite{rebecq2019high} and ETNet~\cite{weng2021event}, on event data to get a pseudo-frame, then directly fed such a frame into SAM without pre-training. The second utilized pre-trained E2VID as encoder to close the gap between event and RGB domain and then feed those features into the segmentation network, e.g., EVDistill~\cite{wang2021evdistill}, DTL~\cite{wang2021dual} and ESS~\cite{sun2022ess}. Finally, there are knowledge distillation-based methods which do not explicitly utilize the pre-trained decoder by trying to transfer knowledge from image backbone onto event data, e.g., NGA~\cite{hu2020learning}, LFI~\cite{deng2021learning}.
	
	\begin{table}
		\caption{Comparison of the segmentation performance between our method and other representative approaches based on MVSEC dataset.}
		\vspace{-0.3cm}
		\label{table:MVSEC}
		\centering
		\footnotesize
		\renewcommand{\arraystretch}{1.1}
		\setlength{\tabcolsep}{1.3mm}
		\begin{tabular}{c | c c c c | c c c c}
			\toprule[1.2pt]
			\multirow{2}{*}{Method}  &\multicolumn{4}{c|}{Indoor} &\multicolumn{4}{c}{Outdoor}\\
			\cline{2-9}
			\multicolumn{1}{c|}{} &mP & mR &mIoU &aIoU &mP & mR &mIoU &aIoU \\
			\hline 
			SAM~\cite{kirillov2023segment}    &0.38   &0.66  &0.24	 &0.41 &0.41  &\underline{0.71}  &0.28  &0.51 \\
			E2VID~\cite{rebecq2019high}    &0.48   &0.68  &0.32	 &0.46  &0.64  &0.59  &\underline{0.38}	 &0.52 \\
			ETNet~\cite{weng2021event}    &\underline{0.54}   &0.60 
			&\underline{0.36}	 &0.43  &\textbf{0.71}  &0.52  &\underline{0.38}	 &0.46 \\
			NGA~\cite{hu2020learning}    &0.46   &0.72  &0.32  &\underline{0.49}  &0.41  &\textbf{0.73}  &0.31	 &0.50   \\
			LFI~\cite{deng2021learning} &0.44   &\underline{0.74}   &0.31  &\underline{0.49}  &0.48  &0.70	 &0.32 &0.54  \\ 
			DTL~\cite{wang2021dual}     &0.44   &\underline{0.74}   &0.32  &\underline{0.49}  &0.52  &0.68	 &0.34 &\textbf{0.55}   \\ 
			EVDistill~\cite{wang2021evdistill} &\textbf{0.61}  &0.47  &0.29  &0.34  &0.64  &0.47  &0.29	 &0.38   \\ 
			ESS~\cite{sun2022ess}  &0.42   &\textbf{0.76}  &0.31	 &\underline{0.49}  &0.46  &0.72  &0.32	 &0.53 \\
			\hline
			Ours    &0.53   &0.70  &\textbf{0.38}  &\textbf{0.49}  &\underline{0.64}  &0.67  &\textbf{0.42} &\textbf{0.55}  \\
			\bottomrule[1.2pt]
		\end{tabular}
		\label{Table2}
		\vspace{-0.5cm}
	\end{table}
	
	\subsection{Experimental Results}
	\noindent \textbf{RGBE-SEG.} In the comparative analysis presented in Table~\ref{table:RGBE-SEG}, our method demonstrates a notable enhancement in performance, yielding a $9\%$ improvement in both mIOU and aIoU over existing methodologies. While certain approaches like EvDistill exhibit higher mP, they fall short in terms of mR compared to our method. Intriguingly, the original SAM, without any fine-tuning, attains remarkably competitive results, particularly in simpler scenarios. This observation underscores the potential detriment to performance when large pre-trained models are not appropriately adapted. Among the various methods evaluated, E2VID emerges as the most effective, underscoring the efficacy of transitioning from event to video data in enhancing event-based object segmentation. Contrarily, methods employing the E2VID encoder experience a significant performance decline, likely due to modifications in the original SAM architecture which adversely affect model generalization. A visual comparison of these methods, illustrated in the first row of Fig.~\ref{fig:visual}, further validates our findings. Our method distinctly differentiates between sky and cloud, and offers clearer segmentation of ground and vehicles compared to others. Notably, it even seems to surpass the SAM with image data in certain aspects, reinforcing the viability of using event data for object segmentation.
	
	% As shown in Table., the proposed method shows a significant advancement than other methods with $9\%$ improvement of mIOU and aIoU.  Although some methods, like EvDistill, have higher AP, their AR is much lower than the proposed methods. It shows interesting that the original SAM, even with out any fine-tune, could achieve a quite competitive performance than other methods, especially in easy case. It also shows that without properly modification of large pre-trained models, their performance might be seriously destroyed. Moreover, among different comparing methods, E2VID  achieves the highest performance. It indicates that applying transition of event to video could indeed improve the performance of event-based object segmentation. However, for the methods utilize the E2VID encoder, their performance seriously degraded, which might be largely attributed to the modification of original SAM architecture seriously compromise model generalization ability. The first row in Fig.~\ref{fig:visual}, presents a visual comparison of different methods. Our method could distinguish the sky and cloud. Moreover, the segment of ground and cars are also clear than other methods, even, more reasonable than SAM with image, further validating that using event data is quite reasonable for object segmentation.

	\noindent \textbf{MVSEC.} The segmentation results of MVSEC are presented in Table~\ref{table:MVSEC}. Our method continues to outperform competing techniques, evidencing robust zero-shot capabilities.  It is observed that other event-based segmentation methods, particularly those employing frame reconstruction, exhibited quite improvements. This enhancement may be attributed to the high similarity of the scenes in the dataset, which simplifies the reconstruction process. Furthermore, as depicted in the third row of Fig.~\ref{fig:visual}, the proposed method effectively distinguishes the lane line and car from background. This is further validated by the consistent results obtained when SAM is applied to the image data, thereby demonstrating the successful adaptation of SAM on the event-based object segmentation task.

	\begin{table}
		\caption{Ablation study results of the fine-tuning layers based on RGBE-SEG dataset. All models are evaluated with event frame input. ``w/o Fine-tuning" indicates the original SAM without any retraining, ``Embed" denotes the embedding layers, ``Four MLPs" represents the last MLPs from $\{3,6,9,12\}$ blocks , ``Four Blocks" also indicates those blocks, ``All MLPs" represents the last MLPs from all the transformer layers, ``All Blocks" denotes all the blocks contained in ViT layers. \underline{Underline} for the adopt strategy (\textbf{our baseline}). Note that all methods are w/o weighted regularization.}
		% \vspace{-0.3cm}
		\centering
		\footnotesize
		\renewcommand{\arraystretch}{1.1}
		\setlength{\tabcolsep}{1.3mm}
		\begin{tabular}{c | c|c c c c}
			\toprule[1.2pt]
			Fine-tuning Layers  & Trainable \#Param &mP & mR &mIoU &aIoU \\
			\hline
			w/o Fine-tuning    &- &0.39   &0.73  &0.26 &0.43  \\
			Embed             &0.6M ($0.7\%$) &0.46   &0.73  &0.32  &0.47  \\
			\underline{Embed + Four MLPs}   &19.5M ($22.3\%$)&0.52   &0.73  &0.37  &0.53     \\
			Embed + Four Blocks   &29.0M ($33.2\%$)&0.51   &0.68   &0.32  &0.47    \\ 
			Embed + All MLPs    &57.3M ($65.7\%$)&0.48   &0.69  &0.30  &0.45     \\
			Embed + All Blocks &87.3M (100$\%$)&0.51  & 0.63 & 0.29 &  0.41  \\
			% Embed + All Blocks  &\multirow{ 2}{*}{0.51}   &\multirow{ 2}{*}{0.63}   &\multirow{ 2}{*}{0.29}  &\multirow{ 2}{*}{0.41}    \\
			% (Fully Fine-tuning)  &   &   &  &    \\ 
			\bottomrule[1.2pt]
		\end{tabular}
		\label{Table:AblatLayer}
		\vspace{-0.6cm}
	\end{table}

	\begin{table}
		\caption{Ablation study results of the mixing tokens and weighted adaptation schemes based on RGBE-SEG dataset, where the first row is a baseline finetuned with only multi-layer feature alignment as $3^{rd}$ row of Table~\ref{Table:AblatLayer},  ``Token Mixing" means to mix RGB image tokens with event tokens for facilitating adaptation, ``$\widetilde{\mathbf{H}}^{(s)}_{E}$" indicates to calculate embedding weights via aggregating event (student) attention matrices, ``$\widetilde{\mathbf{H}}^{(s)}_{M}$" indicates to calculate embedding weights via aggregating image (teacher) attention matrices, ``single $\widetilde{\mathbf{H}}^{(s)}_{M}$" represents to only utilize one layer attention matrix to calculate regularization weights.}
		% \vspace{-0.3cm}
		\centering
		\footnotesize
		\renewcommand{\arraystretch}{1.1}
		\setlength{\tabcolsep}{1.3mm}
		\resizebox{0.46\textwidth}{!}
		{
		\begin{tabular}{ c | c c c c| c c c c}
		\toprule[1.2pt]
		No.	 &Token Mixing &$\widetilde{\mathbf{H}}^{(s)}_{E}$ &$\widetilde{\mathbf{H}}^{(s)}_{M}$ &single $\widetilde{\mathbf{H}}^{(s)}_{M}$ &mP & mR &mIoU &aIoU \\
		\hline
		\textbf{a.} &$\times$          & $\times$        &$\times$        &$\times$        &0.52   &0.73  &0.37  &0.53 \\
		\textbf{b.} &$\checkmark$      &$\times$        &$\times$        &$\times$        &0.53   &0.74  &0.38  &0.54 \\
		\textbf{c.} &$\times$          &$\times$        &$\checkmark$     &$\times$    &0.58   &0.70  &0.40  &0.55 \\
		\textbf{d.} &$\checkmark$       &$\checkmark$     &$\times$        &$\times$        &0.41   &0.77  &0.30  &0.49  \\
		\textbf{e.} &$\checkmark$        & $\times$    &$\times$        &$\checkmark$         &0.52   &0.74  &0.37  &0.54     \\
		\textbf{f.} &$\checkmark$       & $\times$        &$\checkmark$     &$\times$    &0.59   &0.71  &0.41  &0.55     \\
		\bottomrule[1.2pt]		
		\end{tabular}
		}
		\label{table:AblationAttn}
		\vspace{-0.7cm}
	\end{table}

	\subsection{Ablation Study}
	\label{Sec:Ablation}
	
	\textbf{Multi-layer Regularization.} As aforementioned, to manage the strong zero-shot ability of SAM in such a cross-modal distillation task, maintaining the original architecture and weights is one of the keys. To verify this point, we conduct ablation studies, showing network performance with re-training different numbers of parameters. As shown in Table~\ref{Table:AblatLayer}, we gradually increase the number of fine-tuning layers in SAM. We could observe a incremental performance improvement at the beginning stage, which indicates that fine-tuning some layers could indeed improve the domain adaptation ability. However, with the retraining parameters continuing to increase, the generalization ability of adapted SAM tends to gradually decline.
	
\begin{figure}
		\centering
		\includegraphics[clip, width=0.45\textwidth]{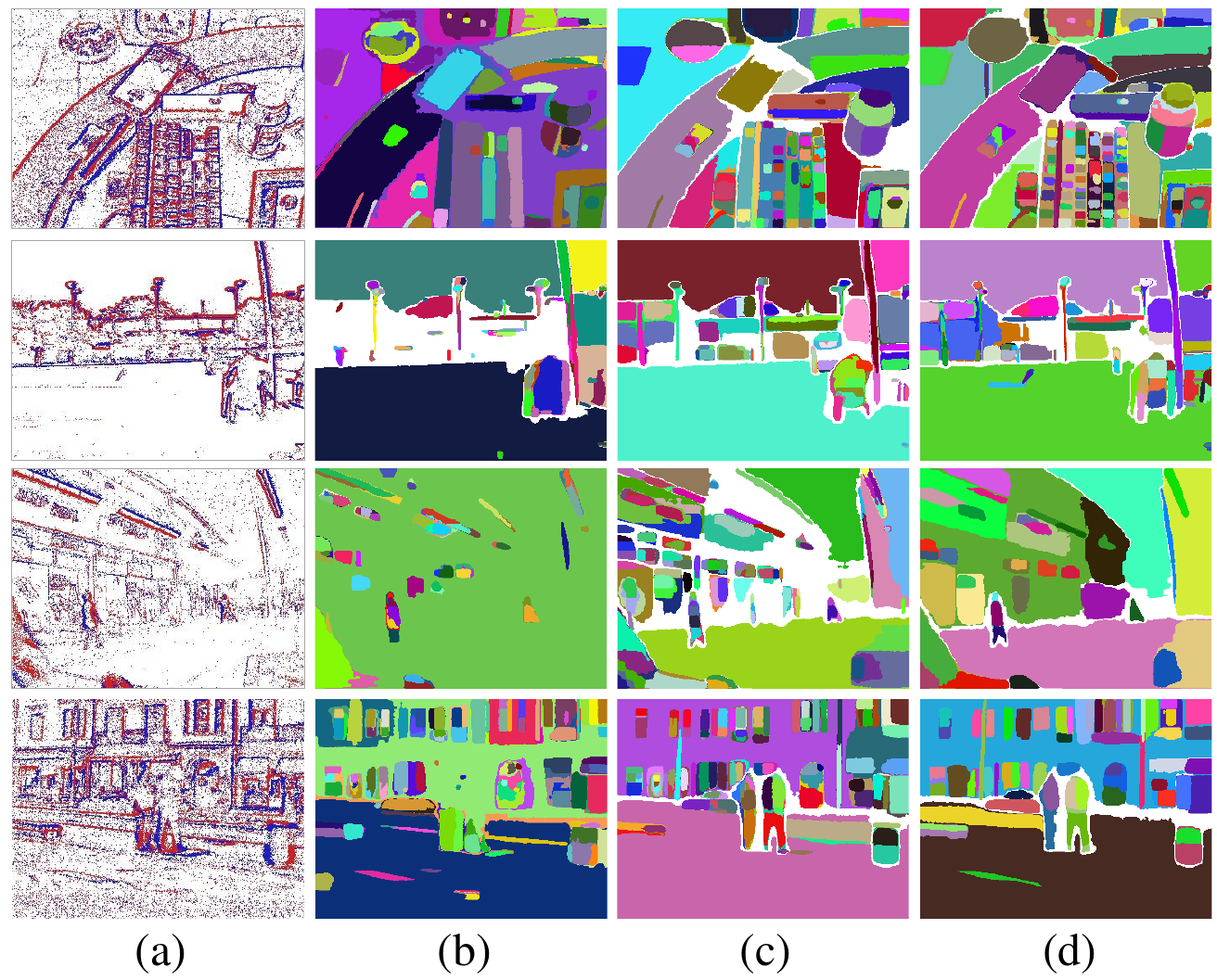}
		% \vspace{-0.2cm}
		\caption{Visual ablation comparison of our designs odawdsdadataset. Each row represents that (a) Event data, (b) Original SAM, (c) Our baseline, and (d) Our method with token mixing and weighted adaptation scheme.}
		\label{fig:Ablation_visual}
		\vspace{-0.7cm}
	\end{figure}

	\noindent\textbf{Attention-aware Weighted Adaptation Scheme.} We approximate the significance of each intermediate embedding by aggregating the attention matrix. To evaluate the effectiveness of such a regularization scheme, we further conduct extensive experiments, as shown in  Table~\ref{table:AblationAttn}. Through comparing \textbf{a.} and \textbf{c.} (or \textbf{b.} and \textbf{f.} ), we could figure out that applying weighted distillation could significantly improve the mIOU with more than $3\%$ and mP of 6$\%$, indicating the effectiveness of the proposed regularization term. The visual comparison of Fig.~\ref{fig:Ablation_visual} further underscores the effectiveness of the proposed strategy. However, such improvement only appears when regularization is derived from the attention matrix of the teacher-transformer, denoted as $\widetilde{\mathbf{H}}^{(s)}_{M}$. Conversely, weighted regularization stemming from $\widetilde{\mathbf{H}}^{(s)}_{E}$ may lead to a deterioration in network performance, evident in \textbf{d)} and \textbf{b)} $7\%$ of mIOU, resulting in a $7\%$ reduction in mIOU. It indicates that weighted cross-modal distillation could indeed play a significant role in the model training. Furthermore, our observations indicate that attention from the student network inadequately captures token embedding significance, thereby resulting in suboptimal performance within this modality.
	
	\noindent\textbf{Token Mixing.} As shown in Table~\ref{table:AblationAttn} \textbf{a.} and \textbf{b.} (\textbf{c.} and \textbf{f.}), through mixing image and event tokens, proposed method achieves  a moderate improvement of $1\%$ consistently on different metrics. Although such improvement is not significant compared with the main contribution of the weighted adaptation scheme, it's still important to help the proposed method manage a leading performance in some datasets, e.g., Table~\ref{table:MVSEC} MVSEC Indoor.
	
	% \noindent\textbf{Gradient \textit{V.S.} Attention Matrix.}

	% \vspace{-0.3cm}
	\section{Conclusions and Discussions}
	In this paper, we have presented a cross-modal adaptation method of SAM for event-centric vision, characterized by an innovative weighted distillation approach. This methodology is inspired by the inherent self-attention mechanism, strategically formulated to reduce discrepancies among key token embeddings, thereby enhancing the adaptation of embeddings from higher-level. Extensive experimental results qualitatively and quantitatively validate the effectiveness of the proposed regularization strategy.
	
	% Besides the aforementioned achievements, their are still some issues and potential improvement for the future work. As current segmentation accuracy is quite low,  we may collect much more data to enrich training set for better adaptation of such event-based SAM. Moreover, integration of this event-based SAM with Large Language Model(LLM) is also choice to directly achieve vision-language integration in event-based vision domain. Finally, such event-adapt SAM could also help other high-level tasks, e.g., object tracking and detection. Note that the SAMs are quite large model, there are still a demands of slimming such network for better adaptation.
	Notwithstanding these significant strides, the study identifies critical areas for future enhancement. To further advance segmentation accuracy, we may need a more voluminous and diverse dataset to refine the adaptation process of the event-based SAM. In addition, exploring the synergy of this specialized SAM with Large Language Models (LLMs) could pioneer new frontiers in achieving seamless integration of vision and language in the context of event-based vision systems. Moreover, the potential application of this adapted SAM extends to other complex tasks such as object tracking and detection. It is imperative to acknowledge that the substantial size of SAMs presents a challenge, highlighting the need for optimization and streamlining of these networks to ensure more efficient adaptation and deployment in diverse scenarios.
	
    {
        \small
        \bibliographystyle{ieeenat_fullname}
        \bibliography{main}
    }
    
    \clearpage
    \includepdf[pages=-]{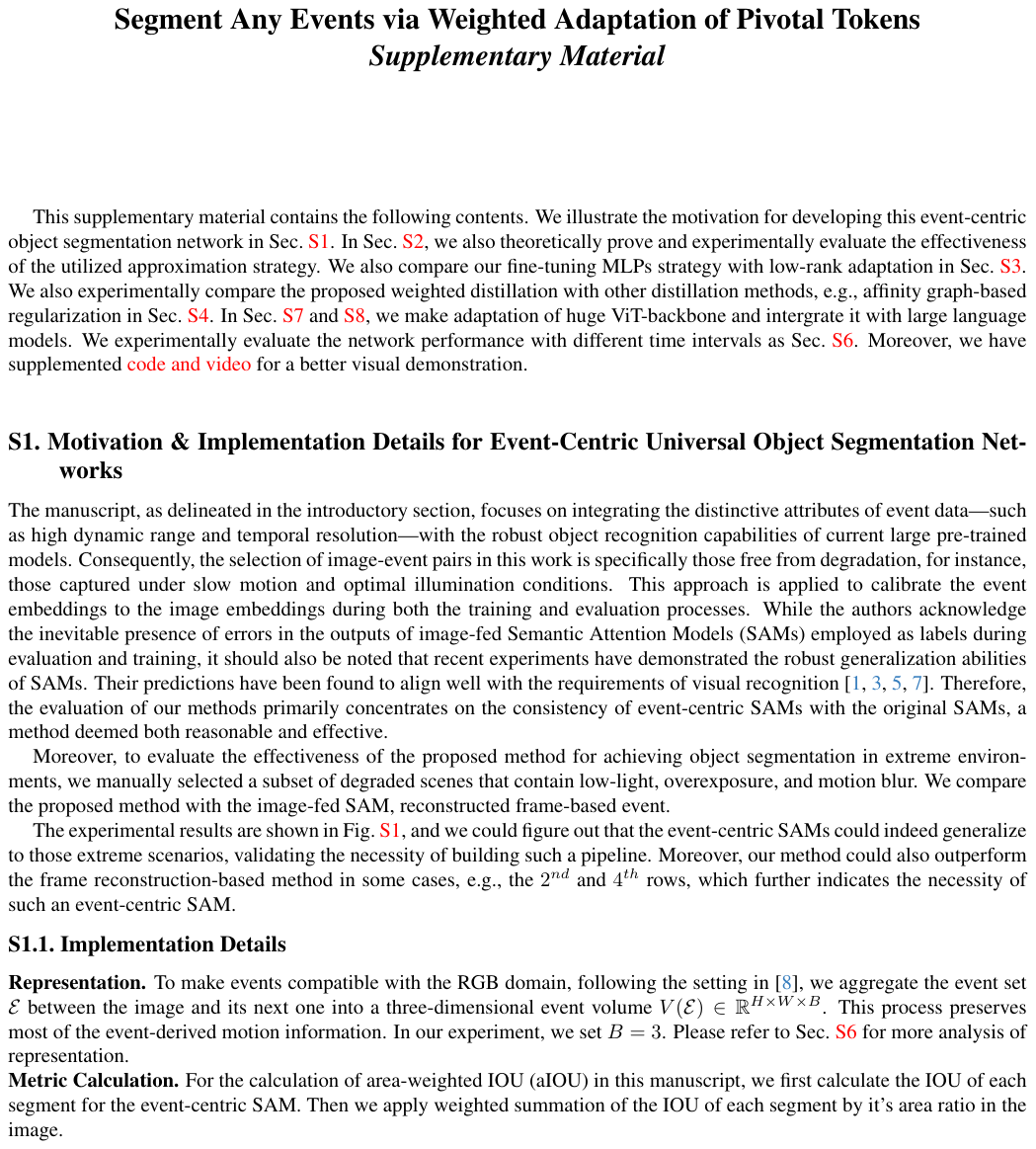}
\end{document}

\maketitle
\thispagestyle{empty}
This supplementary material contains the following contents. We illustrate the motivation for developing this event-centric object segmentation network in Sec.~\ref{SecMotivation}. In Sec.~\ref{SecApprox}, we also theoretically prove and experimentally evaluate the effectiveness of the utilized approximation strategy. We also compare our fine-tuning MLPs strategy with low-rank adaptation in Sec.~\ref{SecLoRA}. We also experimentally compare the proposed weighted distillation with other distillation methods, e.g., affinity graph-based regularization in Sec.~\ref{SecAffinityGraph}. In Sec.~\ref{SecViTH} and~\ref{SecLanguage}, we make adaptation of huge ViT-backbone and intergrate it with large language models. We experimentally evaluate the network performance with different time intervals as Sec.~\ref{SecTime}.
Moreover, we have supplemented \textcolor{red}{code and video} for a better visual demonstration.
% More visual comparisons are shown in Sec.~\ref{SecVisual}.

\vspace{0.5cm}

\section{Motivation \& Implementation Details for Event-Centric Universal Object Segmentation Networks}
\label{SecMotivation}

% As mentioned in the introduction section of the manuscript, we aim to combine the unique characteristics of event data, e.g., high-dynamic range and high-temporal resolution, and strong object recognition ability from the current large pre-trained model. Thus, in the manuscript, we select the image-event pairs without degradation, e.g., under slow motion and well illumination conditions, to calibrate the event embeddings to image embeddings in both the training and evaluation processes. We also acknowledge that there are naturally some errors in the outputs of image-fed SAMs, which are used as labels during our evaluation and training. However, recent experiments have also shown that SAMs actually have strong generalization abilities, and their predictions could well fit the needs of visual recognition~\cite{lai2023lisa,zhang2023sam3d,roy2023sam,deng2023segment}. Thus, the evaluation of those methods is actually focused on the consistency of event-centric SAMs with the original SAMs, which is reasonable and effective.
The manuscript, as delineated in the introductory section, focuses on integrating the distinctive attributes of event data—such as high dynamic range and temporal resolution—with the robust object recognition capabilities of current large pre-trained models. Consequently, the selection of image-event pairs in this work is specifically those free from degradation, for instance, those captured under slow motion and optimal illumination conditions. This approach is applied to calibrate the event embeddings to the image embeddings during both the training and evaluation processes. While the authors acknowledge the inevitable presence of errors in the outputs of image-fed Semantic Attention Models (SAMs) employed as labels during evaluation and training, it should also be noted that recent experiments have demonstrated the robust generalization abilities of SAMs. Their predictions have been found to align well with the requirements of visual recognition~\cite{lai2023lisa,zhang2023sam3d,roy2023sam,deng2023segment}. Therefore, the evaluation of our methods primarily concentrates on the consistency of event-centric SAMs with the original SAMs, a method deemed both reasonable and effective.

Moreover, to evaluate the effectiveness of the proposed method for achieving object segmentation in extreme environments, we manually selected a subset of degraded scenes that contain low-light, overexposure, and motion blur. We compare the proposed method with the image-fed SAM, reconstructed frame-based event.

% we further reconstruct a set of high-quality images using both degraded images and events. Then, we could utilize them as references to calculate the performance of image-centric and event-centric SAMs.

\begin{figure*}[h]
  \centering
    \includegraphics[clip, width=1.0\textwidth]{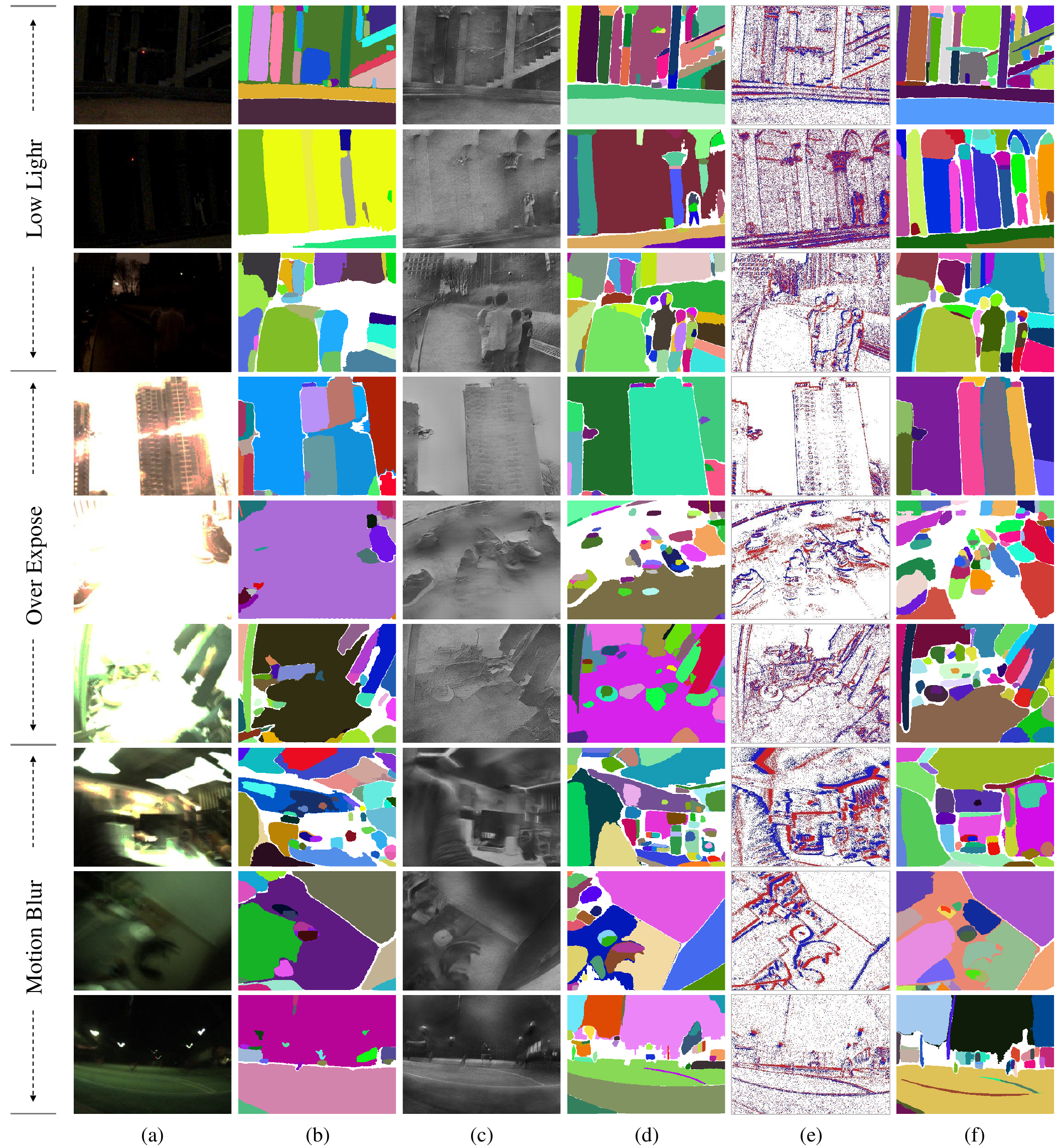}
    \vspace{-0.2cm}
    \caption{Visual comparison of the segmentation results from different degraded scenes, where each column indicate as the following: (a) the degraded image, (b) SAM output w/ image, (c) the reconstructed image by event data (E2VID ~\cite{rebecq2019high}), (d) SAM output w/ reconstructed image, (e) corresponding event data, (f) SAM adapted w/ Ours.}
    \label{fig:Degraded}
\end{figure*}

The experimental results are shown in Fig.~\ref{fig:Degraded}, and we could figure out that the event-centric SAMs could indeed generalize to those extreme scenarios, validating the necessity of building such a pipeline. Moreover, our method could also outperform the frame reconstruction-based method in some cases, e.g., the $2^{nd}$ and $4^{th}$ rows, which further indicates the necessity of such an event-centric SAM.

\subsection{Implementation Details}

\noindent \textbf{Representation.} To make events compatible with the RGB domain, following the setting in \cite{zhu2019unsupervised}, we aggregate the event set $\mathcal{E}$ between the image and its next one into a three-dimensional event volume $V(\mathcal{E}) \in \mathbb{R}^{ H \times W \times B}$. This process preserves most of the event-derived motion information. In our experiment, we set $B = 3$. Please refer to Sec.~\ref{SecTime} for more analysis of representation.

\noindent \textbf{Metric Calculation.} For the calculation of area-weighted IOU (aIOU) in this manuscript, we first calculate the IOU of each segment for the event-centric SAM. Then we apply weighted summation of the IOU of each segment by it's area ratio in the image.
% \begin{equation}
%     \mathbf{S}_{aIOU} = \sum_{j=0}^{L}\tau \mathbf{S}_{mIOU},
% \end{equation}
% where $\tau$

\clearpage
\section{Approximation of Regularization Weights \texorpdfstring{$\widehat{\mathbf{H}}^{(s)}$}{TEXT}}
\label{SecApprox}
In the manuscript, we formulate the transition process as follows: 

\vspace{-0.3cm}
\begin{equation}
\label{Eq:coverge}
   \quad \mathbf{H}^{(s)}= \prod_{i=s}^{n} [\alpha_i\mathbf{P}^{(i)} + (1-\alpha_i)\mathbf{I}], 
\vspace{-0.1cm}
\end{equation}
where $\mathbf{H}^{(s)}\in \mathbb{R}^{k_s \times k_{n+1} }$ symbolizes a comprehensive information transition matrix from a specific $s^{th}$ layer to the terminal $n^{th}$ layer. Moreover, the scalar $\alpha_i$ signifies the scaling influence of the normalization layers and MLPs. Its results are shown as follows:
\vspace{-0.3cm}
\begin{equation}
    \begin{split}
   \quad \mathbf{H}^{(s)}= &(\alpha_s\alpha_{s+1}\cdots\alpha_{n})\mathbf{P}^{(s)}\times\mathbf{P}^{(s+1)}\cdots\times\mathbf{P}^{(n)}+((1-\alpha_{s})\alpha_{s+1}\cdots\alpha_{n})\mathbf{I}\times\mathbf{P}^{(s+1)}\times\cdots\times\mathbf{P}^{(n)}+\\
   &(\alpha_{s}\alpha_{s+1}\cdots(1-\alpha_{n}))\mathbf{P}^{(s)}\times\mathbf{P}^{(s+1)}\times\cdots\times\mathbf{I}+((1-\alpha_{s})(1-\alpha_{s+1})\cdots\alpha_{n})\mathbf{I}\times\mathbf{I}\times\cdots\times\mathbf{P}^{(n)}+\\
   &\cdots  + ((1-\alpha_{s})(1-\alpha_{s+1})\cdots (1-\alpha_{n}))\mathbf{I}\times\mathbf{I}\times\cdots\times\mathbf{I}.
    \end{split}
\vspace{-0.1cm}
\end{equation}

To thoroughly investigate such information flow, we take it as a transition process in markov chain. Here we define each $\mathbf{P}^{(i)}$ as a transition matrix $\mathbf{Q}^{(i)}$ from state $i$ to $i+1$. If such the transition matrix has the following characteristics, the Markov chain would converge to a distribution.

\begin{equation}
\label{Eq:covg}
    \lim_{i \rightarrow +\infty} \mathbf{Q}^{(s)} * \mathbf{Q}^{(s+1)} * \cdots \mathbf{Q}^{(i)} \rightarrow \mathbf{M},
\end{equation}

\noindent where $\mathbf{M}$ indicates the limitation of a series of matrix products. We assume that the attention matrix $\mathbf{P}^{(i)}$ has such convergence properties as Eq.~\ref{Eq:covg} (as experimentally evaluated in Sec.~\ref{SecCovergence}). Thus, we have

\begin{equation}
\label{Eq:covg}
    \|\mathbf{P}^{(s)} * \mathbf{P}^{(s+1)} * \cdots \mathbf{P}^{(i)} - \mathbf{M} \|_2<\delta, \quad \|\mathbf{P}^{(s)} * \mathbf{P}^{(s+1)} * \cdots \mathbf{P}^{(i+1)} - \mathbf{M} \|_2<\delta
\end{equation}
where $\delta$ is a small scalar. Through triangle inequality, we have
\begin{equation}
\label{Eq:covg}
    \|\mathbf{P}^{(s)} \times \mathbf{P}^{(s+1)} \times \cdots \mathbf{P}^{(i)} - \mathbf{P}^{(s)} \times \mathbf{P}^{(s+1)} \times \cdots \mathbf{P}^{(i+1)} \|_2\leq2\delta,
\end{equation}
Moreover, with sub-multiplicative property of matrix norms, we have 
\begin{equation}
\label{Eq:covg}
    \begin{split}
    \|\mathbf{P}^{(s)} \times \mathbf{P}^{(s+1)} \times \cdots \times \mathbf{P}^{(i)} \times \mathbf{P}^{(i+2)}\times \cdots \times \mathbf{P}^{(n)}  - \mathbf{P}^{(s)} \times \mathbf{P}^{(s+1)} \times \cdots \mathbf{P}^{(i+1)}\times \mathbf{P}^{(i+2)}\times \cdots \times \mathbf{P}^{(n)} \|_2 \leq  \\\|\mathbf{P}^{(s)} \times \mathbf{P}^{(s+1)} \times \cdots \mathbf{P}^{(i)} - \mathbf{P}^{(s)} \times \mathbf{P}^{(s+1)} \times \cdots \mathbf{P}^{(i+1)} \|_2 \cdot \| \mathbf{P}^{(i+2)}\times \cdots \times \mathbf{P}^{(n)}\|_2 \leq 2k\delta,
    \end{split}
\end{equation}
where the $\| \mathbf{P}^{(i+2)}\times \cdots \times \mathbf{P}^{(n)}\|_2 \leq k$, where k is not a large number  due to they act as transition matrices. Thus, randomly dropping several intermediate matrices would not greatly influence the transition process. Then we could approximate each term except the last term with $\prod_{i=s}^{n} \mathbf{P}^{(i)}$. It results in 

\begin{equation}
    \label{Eq:approx}
    \begin{split}
   \quad \mathbf{H}^{(s)}&\approx (\alpha_s\alpha_{s+1}\cdots\alpha_{n})\prod_{i=s}^{n} \mathbf{P}^{(i)}+((1-\alpha_{s})\alpha_{s+1}\cdots\alpha_{n})\prod_{i=s}^{n} \mathbf{P}^{(i)}+\cdots + ((1-\alpha_{s})(1-\alpha_{s+1})\cdots (1-\alpha_{n}))\mathbf{I}\\
   &\approx \beta\prod_{i=s}^{n} \mathbf{P}^{(i)} + (1-\beta)\mathbf{I},
    \end{split}
\vspace{-0.1cm}
\end{equation}
where we do not enforce $\beta = (\alpha_s\alpha_{s+1}\cdots\alpha_{n}) + ((1-\alpha_{s})\alpha_{s+1}\cdots\alpha_{n}) + \cdots $ and $(1-\beta) = ((1-\alpha_{s})(1-\alpha_{s+1})\cdots (1-\alpha_{n}))$. Since we expect through coordinate those parameters, it could compensate for some errors in the approximation process.

% To properly select the value of  hyper-parameter $\beta$, we have conducted extensive experiments, as shown in Table.~\ref{table:hyper}, where we can see that the setting with $\beta=0.5$ is with the best performance than others, e.g., 0, 0.25, 0.75, 1.
In this section, we conduct an experimental evaluation to assess the impact of varying values of the hyperparameter $\beta$. The results are presented in Table~\ref{table:hyperbeta}, where we systematically increase $\beta$ from 0 to 1 in increments of 0.25, while keeping other conditions constant.

We observe that the network performance exhibits a gradual improvement, starting from 0.38 and reaching a saturation point at 0.41. As $\beta$ increases, the performance of the network initially rises, but beyond a certain point, further increases in $\beta$ lead to a decline in network performance. Similar results are also observed on the MVSEC dataset. Thus, we select $\beta$ as 0.5.

% \section{Ablation studies of hyperparamter \texorpdfstring{$\beta$}{TEXT}}
% We conduct different experiments 

\begin{table}[h]
	\caption{Ablation study results of the scaling factor $\beta$ (see Eq.~\ref{Eq:approx}) based on RGBE-SEG and MVSEC datasets. Note that all methods are w/ the token mixing scheme.}
 \label{table:hyperbeta}
	\centering
	\footnotesize
	\renewcommand{\arraystretch}{1.3}
	\setlength{\tabcolsep}{1.3mm}
	\begin{tabular}{c | c c c c | c c c c}
		\toprule[1.2pt]
		\multirow{2}{*}{$\beta$}  &\multicolumn{4}{c|}{RGBE-SEG} &\multicolumn{4}{c}{MVSEC}\\
		\cline{2-9}
		\multicolumn{1}{c|}{} &mP & mR &mIoU &aIoU &mP & mR &mIoU &aIoU \\
        \hline 
        0.0    &0.53   &\textbf{0.74}  &0.38	 &0.54 &0.53  &\textbf{0.71}  &0.38  &\underline{0.52} \\
	0.25    &0.57   &0.72  &0.39	 &\underline{0.55}  &0.58  &0.69  &0.39	 &\underline{0.52} \\
        0.5    &0.59   &0.71 &\textbf{0.41}	 &\textbf{0.55}  &0.59  &0.69  &\textbf{0.40}	 &\textbf{0.52} \\
        0.75    &\textbf{0.61}   &0.69  &0.40  &\underline{0.55}  &\textbf{0.61}  &0.68  &0.40	 &\underline{0.52}   \\
        1.0 &0.55   &0.72   &0.39  &\underline{0.55}  &0.57  &0.70	 &0.39 &\underline{0.52}  \\ 
		\bottomrule[1.2pt]
	\end{tabular}
	\label{Table1}
\end{table}

\subsection{Experimental Validation of the Assumption of Attention-transition} 
\label{SecCovergence}
Our approximation is based on the assumption that the product of multiple layers' attention matrices will eventually converge to a matrix $\mathbf{M}$. In order to validate this assumption and examine the convergence properties, we calculate the L2 norm of the following quantities:

\begin{equation}
\label{Eq:covg}
    \mathcal{P}^{i} = \norm{\prod_{t=0}^{i}\mathbf{P}^{(t)}- \prod_{t=0}^{n}\mathbf{P}^{(t)}}_2 ,
\end{equation}
where $n$ indicates a total number of transition matrices in a ViT backbone. 

\begin{figure*}[h]
  \centering
    \includegraphics[clip, width=0.5\textwidth]{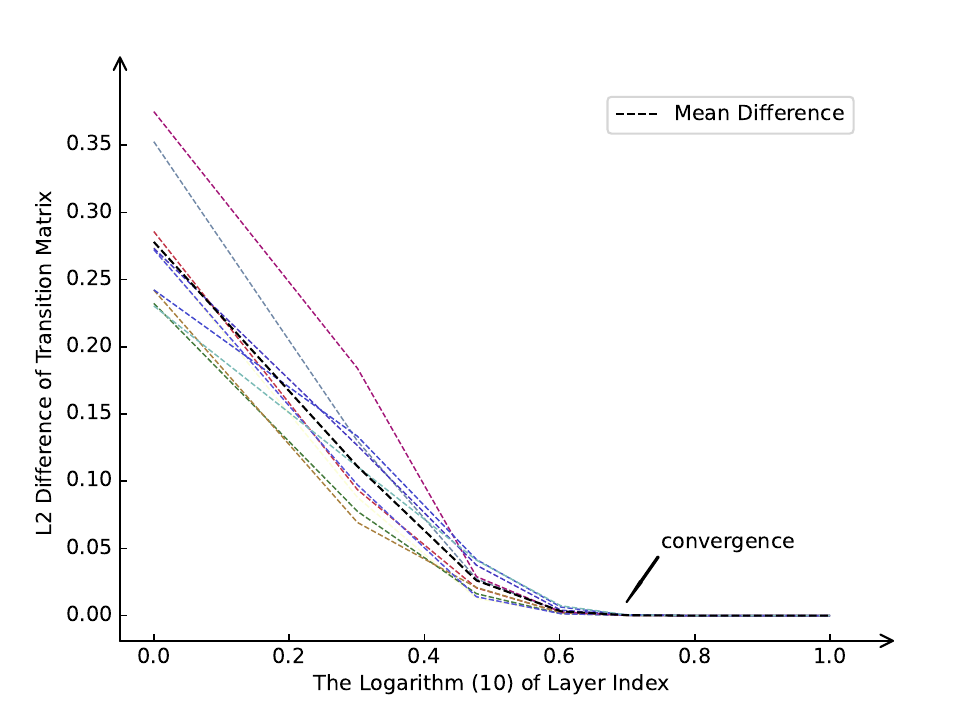}
    \vspace{-0.2cm}
  \caption{Illustration of the trends of $\mathcal{P}^{i}$ with variation of layer index $i$}
    \label{fig:Matrix_Diff}
% \vspace{-0.6cm}
\end{figure*}

The convergence of our approximation can be observed in Fig.~\ref{fig:Matrix_Diff}, where the values of $\mathcal{P}^{i}$ decrease as $i$ increases. This trend indicates that the product $\prod_{t=0}^{i}\mathbf{P}^{(t)}$ gradually approaches zero, supporting the notion that it converges to the desired matrix $\textbf{M}$ as demonstrated in Eq.~\ref{Eq:coverge}. The observed convergence provides validation for the correctness of our approximation. We also acknowledge that unrolling $\mathbf{H}^{(s)}$ leads to $2^i$ terms, each of which, when approximated, could contribute to a quantification error. To limit the cumulative error growth, the approach taken in this work is to consider only the subsequent three layers for each layer.

\subsection{Necessity of such approximation}

To validate the effectiveness of our approximation strategy, we also conduct extensive experiments do directly adopt Eq.~\ref{Eq:coverge} to calculate token significance. To achieve this, we manually set all alpha as the same. Due to that for some $\alpha$ the training losses do not deline, we only report the alpha greater than 0.9. The results indicate that without the properly setting the each $\alpha$, the weighted KD loss even degrades the network performance. 
% \section{More Implementation Details}
% \subsection{Hyperparameter $\beta$}

\begin{table}[h]
	\caption{Ablation study results of the scaling factor $\alpha$ (see Eq.~\ref{Eq:coverge}) based on RGBE-SEG and MVSEC datasets. For the relatively small $\alpha$, e.g., 0.8 , the training loss indeed does not decline. Thus, we only report the experimental results with relatively large   $\alpha$. Note that all methods are w/ token mixing.}
 \label{table:hyperalpha}
	\centering
	\footnotesize
	\renewcommand{\arraystretch}{1.3}
	\setlength{\tabcolsep}{1.3mm}
	\begin{tabular}{c | c c c c | c c c c}
		\toprule[1.2pt]
		\multirow{2}{*}{$\alpha$}  &\multicolumn{4}{c|}{RGBE-SEG} &\multicolumn{4}{c}{MVSEC}\\
		\cline{2-9}
		\multicolumn{1}{c|}{} &mP & mR &mIoU &aIoU &mP & mR &mIoU &aIoU \\
        \hline 
        0.99    &0.39   &0.80  &0.31 &0.50 &0.41  &0.80  &0.31  &0.52 \\
	0.95    &0.38   &0.81  &0.30 &0.50 &0.41  &0.80  &0.31  &0.51 \\
        0.90    &0.38   &0.79  &0.29 &0.49 &0.40  &0.80  &0.31  &0.51 \\
        Our baseline w/ token mixing &0.53   &0.74  &0.38 &0.54 &0.53  &0.71  &0.38  &0.52 \\
		\bottomrule[1.2pt]
	\end{tabular}
	\label{Table1}
\end{table}

\section{Comparison with Low Rank Adaptation (LoRA)~\cite{hu2021lora}}
\label{SecLoRA}
In the proposed method, we directly finetune the MLPs, instead of using some popular model adaptation tricks, e.g., LoRA. In this section, we conduct experiments to investigate our performance against LoRA. As shown in Table.~\ref{table:lora}, we try finetune the SAM with different methods. All methods are trained on the same dataset (RGBE-SEG training) and tested on RGBE-SEG and MVSEC, respectively. Note that the training configuration of those methods are exactly same, except the trainable parameters. The experimental results show that applying LoRA to the SAM adaptation cannot achieve considerable performance.

\begin{table}[h]
	\caption{Comparison of the segmentation performance between LoRA and our fine-tuning method based on RGBE-SEG and MVSEC datasets. And we have set up a series of ranks (r) of LoRA to fully explore its adaption effect. Note that all methods are w/o the token mixing and weighted regularization.}
 % \vspace{-0.3cm}
 \label{table:lora}
	\centering
	\footnotesize
	\renewcommand{\arraystretch}{1.3}
	\setlength{\tabcolsep}{1.3mm}
	\begin{tabular}{c | c | c c c c | c c c c}
		\toprule[1.2pt]
		\multirow{2}{*}{Fine-tuning Method} &\multirow{2}{*}{Trainable \#Param}  &\multicolumn{4}{c|}{RGBE-SEG} &\multicolumn{4}{c}{MVSEC}\\
		\cline{3-10}
		\multicolumn{1}{c|}{} & &mP & mR &mIoU &aIoU &mP & mR &mIoU &aIoU \\
        \hline
        w/o Fine-tuning    &-  &0.39 &0.73  &0.26 &0.43  &0.40  &0.68   &0.26  &0.46  \\
        LoRA(Embed + Four MLPs, r=16)   &1.1M &0.39   &\textbf{0.76}  &0.28	 &0.44  &0.36  &0.75  &0.25	 &0.45  \\
        LoRA(Embed + Four MLPs, r=64)   &2.6M &0.40   &\underline{0.76}  &0.28	 &0.44  &0.36  &0.75  &0.25	 &0.45  \\
        LoRA(Embed + Four MLPs, r=256)   &8.5M &\underline{0.41}   &0.75  &0.29	 &0.45  &0.37  &0.74  &0.26	 &0.45  \\
	LoRA(Embed + All Blocks, r=16)   &2.9M &0.40   &0.75 &0.27	 &0.43  &0.35  &\textbf{0.76}  &0.24	 &0.44  \\
        LoRA(Embed + All Blocks, r=64)   &10.0M &\underline{0.41}   &0.74 &0.28	 &0.44  &0.33  &\underline{0.76}  &0.24	 &0.45 \\
        LoRA(Embed + All Blocks, r=256)  &38.4M  &0.40   &0.74  &0.28  &0.44  &0.34  &\underline{0.76}  &0.24	 &0.45   \\
        Our baseline &29.0M  &\textbf{0.52}   &0.73   &\textbf{0.37}  &\textbf{0.53}  &\textbf{0.53}  &0.69 &\textbf{0.37} &\textbf{0.52}  \\ 
		\bottomrule[1.2pt]
	\end{tabular}
	\label{Table1}
\end{table}

\vspace{-0.5cm}
\section{Comparison with Affinity Graph KD}
\label{SecAffinityGraph}
\begin{figure*}[h]
  \centering
    \includegraphics[clip, width=\textwidth]{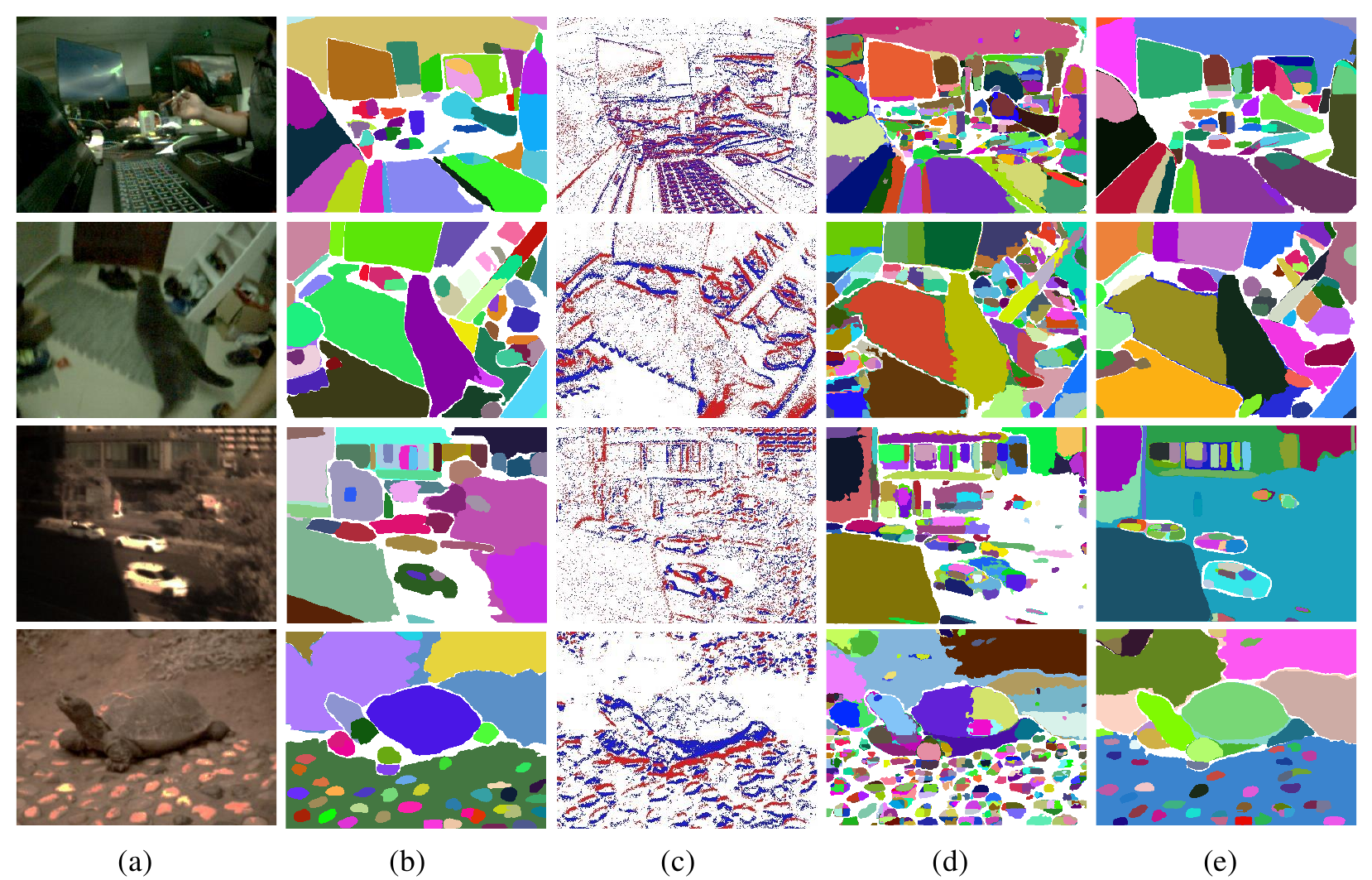}
    \vspace{-0.2cm}
  \caption{Visual comparison of the segmentation results from different KD manners, where each figure indicate as the following: (a) the reference image, (b) SAM output w/ image, (c) corresponding event data, (d) SAM adapted w/ affinity graph, (e) SAM adapted w/ Ours.}
    \label{fig:KDManner}
\end{figure*}

We also consider a comparison method for embedding KD, namely affinity graph-based KD~\cite{wang2021dual}. In our evaluation, we focus on modifying the loss function to explore different knowledge distillation approaches while keeping other settings consistent. The experimental results, presented in Table~\ref{table:KDManner}, demonstrate that our proposed methods achieve superior improvements in terms of mIOU and maintain higher aIOU values. In contrast, the affinity graph-based KD approach exhibits a significant decrease in aIOU, indicating difficulties in achieving accurate segmentation over large areas. This observation is further supported by Fig.~\ref{fig:KDManner}-(d), where the affinity graph-based method only successfully segments small regions and struggles to capture the global contour of objects.
\begin{table}[h]
	\caption{Comparison of the segmentation performance between affinity graph and our KD scheme based on RGBE-SEG and MVSEC datasets. Note that all methods are w/ the token mixing scheme.}
 \label{table:KDManner}
	\centering
	\footnotesize
	\renewcommand{\arraystretch}{1.3}
	\setlength{\tabcolsep}{1.3mm}
	\begin{tabular}{c | c c c c | c c c c}
		\toprule[1.2pt]
		\multirow{2}{*}{KD Scheme}  &\multicolumn{4}{c|}{RGBE-SEG} &\multicolumn{4}{c}{MVSEC}\\
		\cline{2-9}
		\multicolumn{1}{c|}{} &mP & mR &mIoU &aIoU &mP & mR &mIoU &aIoU \\
        \hline 
        Our baseline w/ token mixing    &0.53 &0.74 &0.38 &0.54  &0.53  &0.71  &0.38  &0.52  \\
        \hline 
        Affinity Graph    &\textbf{0.67}   &0.61  &0.40	 &0.49  &\textbf{0.61}  &0.63  &0.38  &0.46  \\
        Incremental   &\textcolor{red}{+0.14}   &\textcolor{cyan}{-0.13}  &\textcolor{red}{+0.02}	 &\textcolor{cyan}{-0.05}  &\textcolor{red}{+0.08}  & \textcolor{cyan}{-0.07} &0 & \textcolor{cyan}{-0.06}\\
        \hline
	Ours    &0.59   &\textbf{0.71} &\textbf{0.41}	 &\textbf{0.55}  &0.59  &\textbf{0.69}  &\textbf{0.40}	 &\textbf{0.52} \\
        Incremental    &\textcolor{red}{+0.06}   &\textcolor{cyan}{-0.03}  &\textcolor{red}{+0.06}	 &\textcolor{red}{+0.01}  &\textcolor{red}{+0.05}  & \textcolor{cyan}{-0.02} &\textcolor{red}{0.02} & 0 \\
		\bottomrule[1.2pt]
	\end{tabular}
	\label{Table1}
\end{table}

% \section{Event-centric SAM for Degraded Scenes}
% \label{SecDegraded}

% \begin{table}[h]
% 	\caption{Segmentation metrics on the degraded test dataset.}
%  \label{table:degraded}
% 	\centering
% 	\footnotesize
% 	\renewcommand{\arraystretch}{1.3}
% 	\setlength{\tabcolsep}{1.3mm}
% 	\begin{tabular}{ c | c c c c | c c c c| c c c c | c c c c}
% 		\toprule[1.2pt]
% 		\multirow{2}{*}{Reference Mask}  &\multicolumn{4}{c|}{Low Light} &\multicolumn{4}{c|}{Over Expose} &\multicolumn{4}{c|}{Motion Blur} &\multicolumn{4}{c}{All}\\
% 		\cline{2-17}
% 		\multicolumn{1}{c|}{} &mP & mR &mIoU &aIoU &mP & mR &mIoU &aIoU &mP & mR &mIoU &aIoU &mP & mR &mIoU &aIoU \\
%         \hline 
%         SAM w/ Image   &0.63   &0.73 &0.45 &0.59  &0.49  &0.74  &0.35	 &0.54 &0.52   &0.67 &0.34 &0.56  &0.54   &0.71 &0.38 &0.56  \\
% 	SAM w/ E2VID   &0.64   &0.68 &0.44 &0.57 &0.54   &0.66 &0.37	 &0.59  &0.40  &0.62  &0.26	 &0.50  &0.53   &0.65 &0.36 &0.55 \\  
% 		\bottomrule[1.2pt]
% 	\end{tabular}
% 	\label{Table1}
% \end{table}

% \clearpage
% \section{More Visual Comparisons}
% \label{SecVisual}

% \clearpage

\section{Different Event Representation}
\label{Sec:representation}
We further conduct experiments to compare affect of different representations. Specifically, we change the patch embedding layer in SAM with recurrent layer for better temporal modeling. However, experimental results (as shown in Table~\ref{table:recurrent}) show that after changing even the embeddings layer, the generalization ability of network gets serious degradation.
\begin{table}[h]
	\caption{Comparison of the segmentation metrics between SAM with recurrent and without recurrent modeling for feature embedding.}
 \label{table:recurrent}
	\centering
	\footnotesize
	\renewcommand{\arraystretch}{1.3}
	\setlength{\tabcolsep}{1.3mm}
	\begin{tabular}{ c | c c c c | c c c c}
		\toprule[1.2pt]
		\multirow{2}{*}{Embedding Layer}   &\multicolumn{4}{c|}{RGBE-SEG} &\multicolumn{4}{c}{MVSEC}\\
		\cline{2-9}
		\multicolumn{1}{c|}{}  &mP & mR &mIoU &aIoU &mP & mR &mIoU &aIoU \\
        \hline 
        w recurrent     &0.39   &0.74 &0.27	 &0.45  &0.42  &0.78  &0.32	 &0.49  \\
	w/o recurrent  &0.59   &0.71 &0.41	 &0.55  &0.59  &0.69  &0.40	 &0.52 \\
		\bottomrule[1.2pt]
	\end{tabular}
	\label{Table1}
\end{table}

\section{Time Interval}
\label{SecTime}
We experimentally validate the network performance with different time interval, where Table~\ref{table:interval} lists the experimental results. The adopted 40ms achieves the best performance.
\begin{table}[h]
	\caption{Ablation study results of the time interval based on RGBE-SEG and MVSEC datasets. \underline{Underline} indicates our time interval setting in the paper. Note that all methods are w/ the token mixing and weighted regularization.}
 \label{table:interval}
	\centering
	\footnotesize
	\renewcommand{\arraystretch}{1.3}
	\setlength{\tabcolsep}{1.3mm}
	\begin{tabular}{c | c c c c | c c c c}
		\toprule[1.2pt]
		\multirow{2}{*}{time interval (ms)}  &\multicolumn{4}{c|}{RGBE-SEG} &\multicolumn{4}{c}{MVSEC}\\
		\cline{2-9}
		\multicolumn{1}{c|}{} &mP & mR &mIoU &aIoU &mP & mR &mIoU &aIoU \\
        \hline 
        10    &0.40   &0.79  &0.30	 &0.45 &0.44  &0.74  &0.31  &0.51 \\
	20    &0.52   &0.73  &0.37	 &0.51  &0.53  &0.71  &0.37	 &0.51 \\
        30    &0.56   &0.73 &0.39	 &0.53  &0.57  &0.70  &0.39	 &0.52 \\
        \underline{40}    &0.59   &0.71  &0.41  &0.55  &0.59  &0.69  &0.40	 &0.52   \\
        50 &0.57   &0.74   &0.39  &0.54  &0.59  &0.69	 &0.40 &0.52  \\ 
        60 &0.56   &0.74   &0.38  &0.53  &0.59  &0.68	 &0.39 &0.52  \\ 
		\bottomrule[1.2pt]
	\end{tabular}
\end{table}

\section{Adaptation of SAM with ViT-Huge}
\label{SecViTH}
\begin{figure*}[h]
  \centering
    \includegraphics[clip, width=\textwidth]{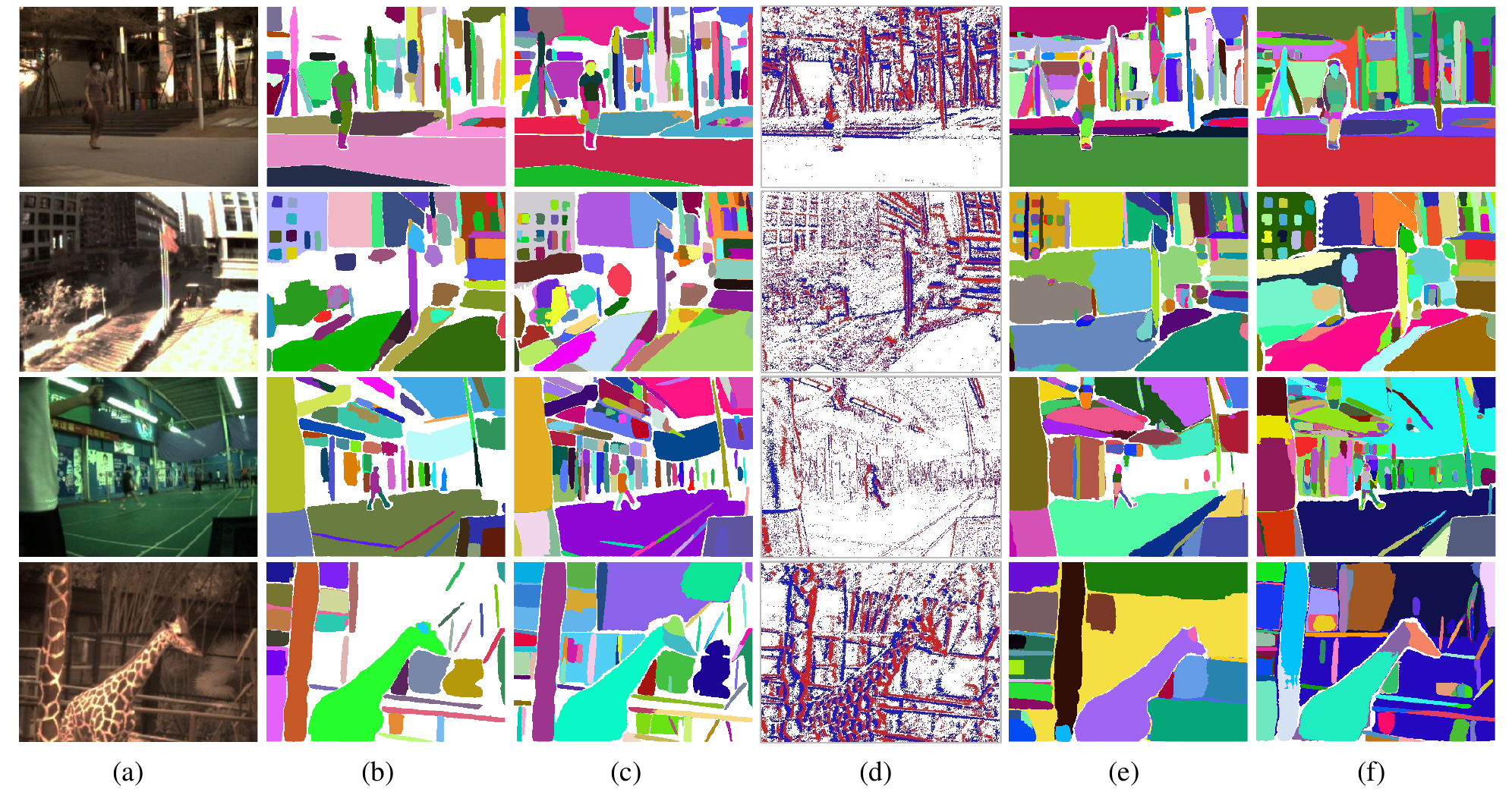}
    \vspace{-0.2cm}
  \caption{Visual comparison of the segmentation results from different SAM encoders, where each figure indicate as the following: (a) the reference image, (b) ViT-B SAM output w/ image, (c) ViT-H SAM output w/ image, (d) corresponding event data, (e) ViT-B SAM output w/ event data, (f) ViT-H SAM output w/ event data.}
    \label{fig:ViT-H}
\end{figure*}

In this section, we further adapt a large SAM with ViT-H as backbone. The experimental results are shown as Fig.~\ref{fig:ViT-H} and table~\ref{table:vit-h}. The ViT-H has stronger modeling ability than ViT-B. Thus, it easy for network ViT-H to approach the results from ViT-B, 

\begin{table}[h]
	\caption{Comparison of the segmentation metrics between SAM with ViT-Base and SAM with ViT-Huge with different reference masks. \underline{Underline} indicates our setting in the paper.}
 \label{table:vit-h}
	\centering
	\footnotesize
	\renewcommand{\arraystretch}{1.3}
	\setlength{\tabcolsep}{1.3mm}
	\begin{tabular}{c | c | c c c c | c c c c}
		\toprule[1.2pt]
		\multirow{2}{*}{Encoder} &\multirow{2}{*}{Reference Mask}  &\multicolumn{4}{c|}{RGBE-SEG} &\multicolumn{4}{c}{MVSEC}\\
		\cline{3-10}
		\multicolumn{1}{c|}{} &  &mP & mR &mIoU &aIoU &mP & mR &mIoU &aIoU \\
        \hline 
        \underline{ViT-B}    &\underline{ViT-B}  &0.59   &0.71 &0.41	 &0.55  &0.59  &0.69  &0.40	 &0.52  \\
	ViT-H    &ViT-H   &0.59   &0.67 &0.39	 &0.53  &0.61  &0.64  &0.39	 &0.53 \\  
        ViT-B    &ViT-H   &0.49   &0.72  &0.35	 &0.51  &0.49  &0.70  &0.34  &0.51  \\
	ViT-H    &ViT-B    &0.66   &0.65  &0.43	 &0.54  &0.69  &0.60  &0.42	 &0.52 \\
		\bottomrule[1.2pt]
	\end{tabular}
	\label{Table1}
\end{table}

\vspace{-0.5cm}
\section{Integration the Proposed Event-centric SAM with Large Language Models}
\label{SecLanguage}
\begin{figure*}[h]
  \centering
    \includegraphics[clip, width=0.7\textwidth]{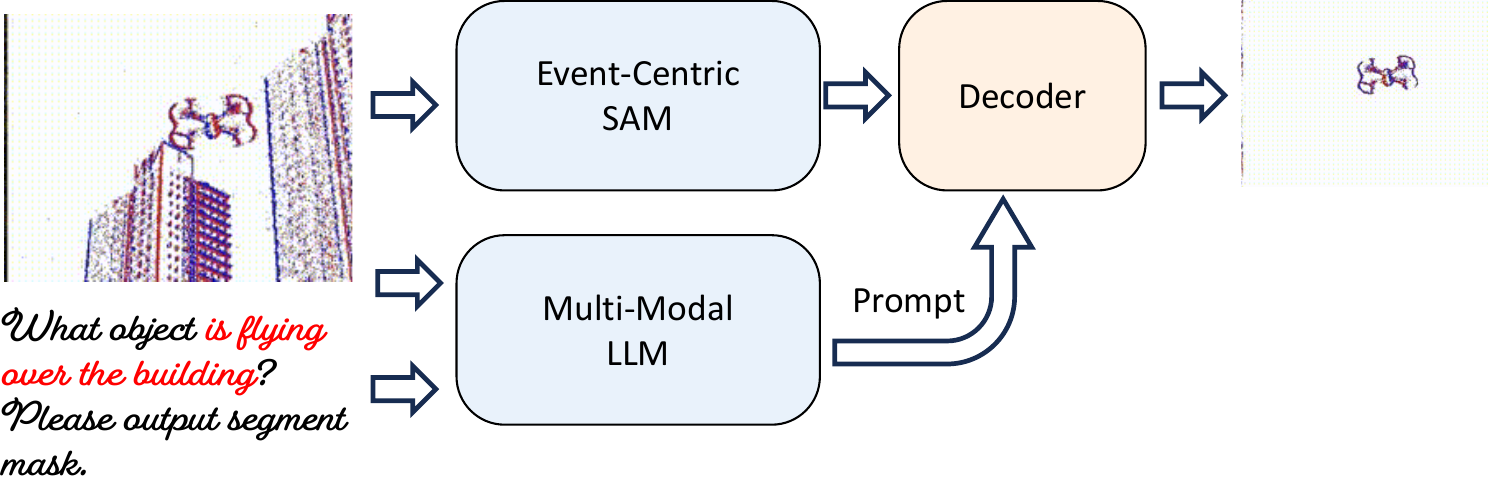}
    \vspace{-0.2cm}
  \caption{Illustration of integration of our event-centric SAM with LLM, where we utilize \textbf{LLaVA-13B-v1-1} as LLM backbone and \textbf{ViT-H} as the backbone of SAM.}
    \label{fig:ViT-H-LLM}
\end{figure*}

\begin{figure*}[h]
  \centering
    \includegraphics[clip, width=0.95\textwidth]{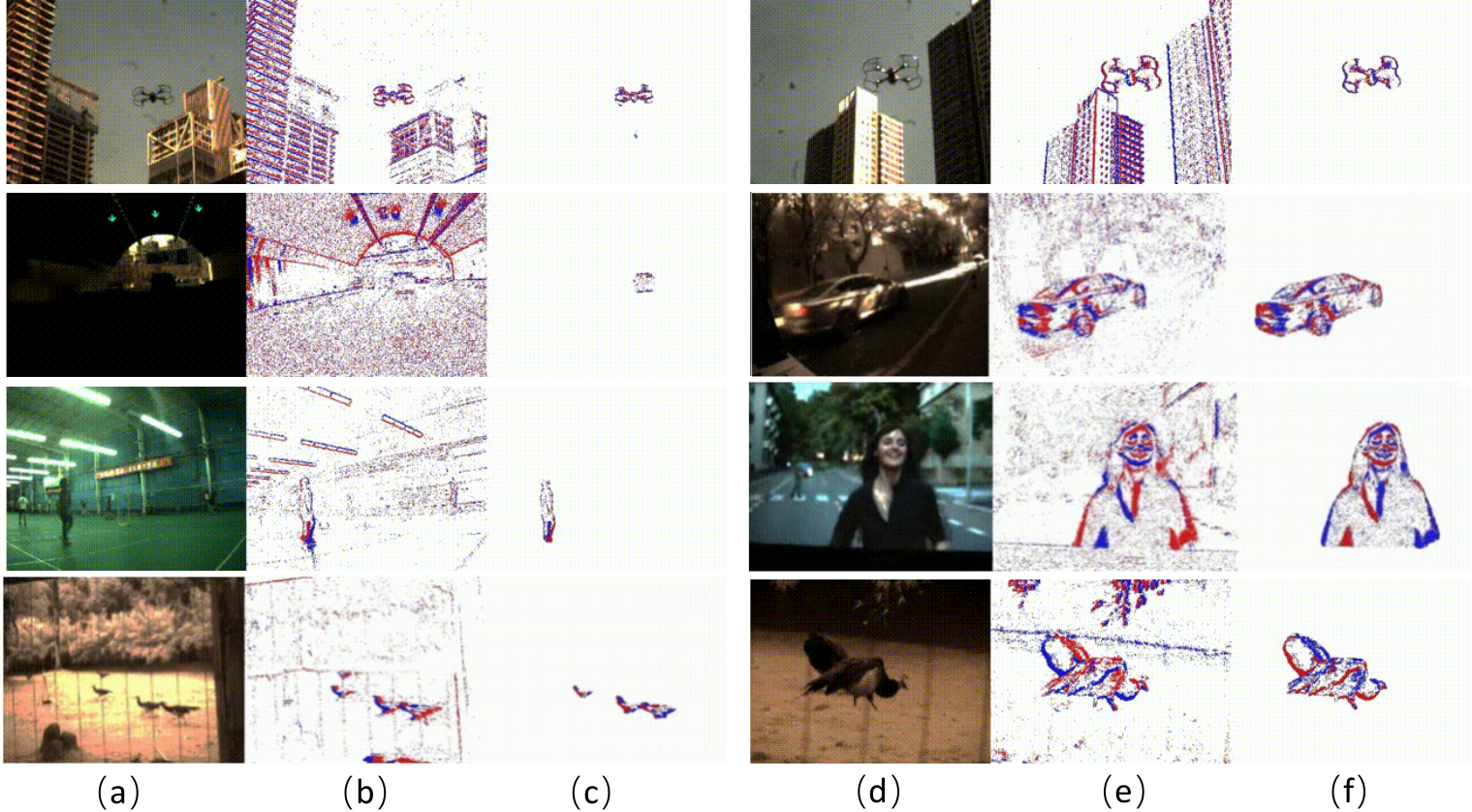}
    \vspace{-0.2cm}
  \caption{Segmentation results of language-prompt event segmentation, where (a) and (d) indicate the RGB images only for visualization not perceived by network, (b) and (e) represents input event, finally (c) and (f) for segmentation results.}
    \label{fig:Lisa-Results}
\end{figure*}

To further validate the strong zero-shot object recognition ability of our event-adapt SAM. We integrate it with a vision-language object segmentation framework LISA~\cite{lai2023lisa}. The framework is shown as Fig.~\ref{fig:ViT-H-LLM}. Through this, we could further unlock the rich semantic inherent in SAM, for interactive universal object segmentation with Event data. Visual segmentation results are shown as Fig.~\ref{fig:Lisa-Results}. Please refer to supplementary video for more visualizations.

% \vspace{-0.5cm}
% \section{Notion of Each Variable}
% We annotate the variable notion of the training pipeline figure as Fig.~\ref{fig:Framework}
% \begin{figure*}[h]
%   \centering
%     \includegraphics[clip, width=0.9\textwidth]{24/Figs/Framework.pdf}
%   \caption{Training workflow for knowledge distillation from pre-trained SAMs to the event domain.}
%     \label{fig:Framework}
% \end{figure*}

% \section{Recurrent Modeling of the Proposed Event-centric SAM}
% \label{SecRecurrent}
% To further validate the sequence memory ability of our event-adapt SAM. We integrate it with a recurrent embedding layer. 

\clearpage
{   \small
    \bibliographystyle{ieeenat_fullname}
    \bibliography{main}}